\documentclass{egpubl}
\usepackage{pg2021}

\SpecialIssuePaper         

\usepackage[T1]{fontenc}
\usepackage{dfadobe}
\usepackage{amsmath}
\usepackage{amssymb}
\usepackage{multirow}
\usepackage{booktabs}
\usepackage{blkarray}
\usepackage{multirow}
\usepackage{tabularx}
\usepackage{bm}
\usepackage{url}
\usepackage{array}
\usepackage{cite}  
\BibtexOrBiblatex
\electronicVersion
\PrintedOrElectronic
\ifpdf \usepackage[pdftex]{graphicx} \pdfcompresslevel=9
\else \usepackage[dvips]{graphicx} \fi

\usepackage{egweblnk}

\title[Diverse Dance Synthesis via Keyframes with Transformer Controllers]%
      {Diverse Dance Synthesis via Keyframes with Transformer Controllers}

\author[Junjun Pan \& Siyuan Wang \& Junxuan Bai \& Ju Dai]
{\parbox{\textwidth}{\centering Junjun Pan$^{1,2}$  
           and Siyuan Wang$^{1}$
           and Junxuan Bai$^{1, 2}$
           and Ju Dai$^{2}$\thanks{The corresponding author: daij@pcl.ac.cn.}
        }
        \\
{\parbox{\textwidth}{\centering $^1$Beihang University, State Key Laboratory of Virtual Reality Technology and Systems, Beijing, China\\
         $^2$Peng Cheng Laboratory, Shenzhen, China
       }
}
}

\begin{document}

\maketitle
\begin{abstract}
Existing keyframe-based motion synthesis mainly focuses on the generation of cyclic actions or short-term motion, 
such as walking, running, and transitions between close postures.
However, these methods will significantly degrade the naturalness and diversity of the synthesized motion 
when dealing with complex and impromptu movements, \emph{e.g.}, dance performance and martial arts.
In addition, current research lacks fine-grained control over the generated motion, which is essential 
for intelligent human-computer interaction and animation creation.
In this paper, we propose a novel keyframe-based motion generation network based on multiple 
constraints, which can achieve diverse dance synthesis via learned knowledge.
Specifically, the algorithm is mainly formulated based on the recurrent neural network (RNN) and the Transformer architecture.
The backbone of our network is a hierarchical RNN module composed of two long short-term memory (LSTM) units, 
in which the first LSTM is utilized to embed the posture information of the historical frames into a latent space, 
and the second one is employed to predict the human posture for the next frame. 
Moreover, our framework contains two Transformer-based controllers, which are used to model the constraints 
of the root trajectory and the velocity factor respectively, so as to better utilize the temporal context of the 
frames and achieve fine-grained motion control.
We verify the proposed approach on a dance dataset containing a wide range of contemporary dance.  
The results of three quantitative analyses validate the superiority of our algorithm. 
The video and qualitative experimental results demonstrate that the complex motion sequences generated by 
our algorithm can achieve diverse and smooth motion transitions between keyframes, even for long-term synthesis. 

\ccsdesc[500]{Computing methodologies~Motion processing}
\ccsdesc[500]{Computing methodologies~Motion capture}

%

\printccsdesc   
\end{abstract}

\begin{figure*}[htbp]
  \centering
  \includegraphics[width=1.0\linewidth, height=.22\linewidth]{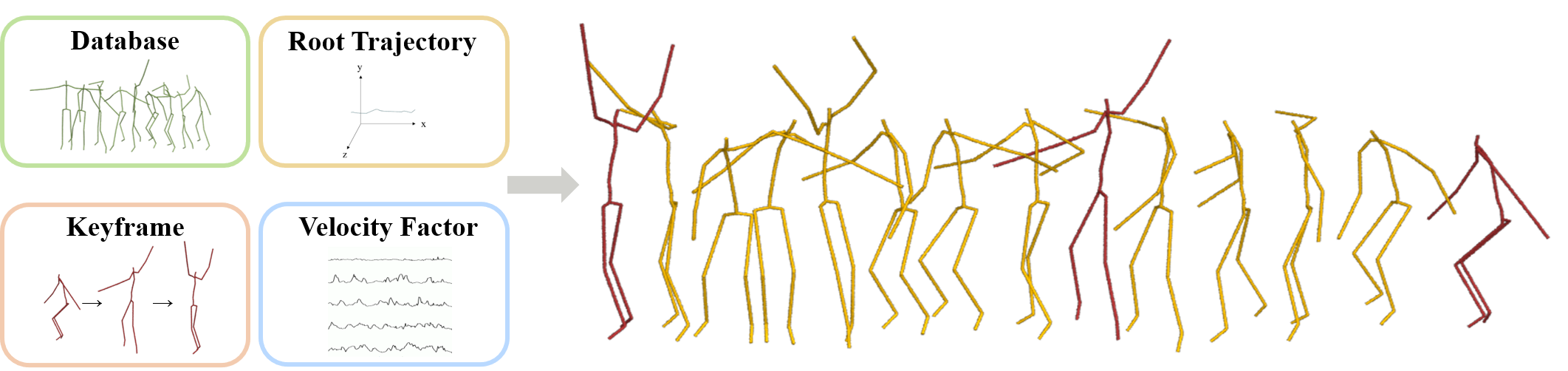}
  \caption{\label{fig:demo}%
           Dance generation from keyframes conditioned on the root trajectory and the velocity factor constraints. 
           The postures marked in red are the keyframes, and the postures marked in yellow are the synthesized frames.}
\vspace{-3mm}
\end{figure*}

\section{Introduction}
Character animation is one of the essential research topics in computer graphics.
Animators usually use motion capture systems or keyframe-based techniques to obtain high-quality animation data. 
However, massive editing and modification still need to be executed for the final animation production, 
which is quite tedious and time-consuming. 
With the development of deep learning techniques, scholars have made significant efforts to simplify the motion 
synthesis process. A prevailing trend is to utilize deep neural networks to generate natural and diverse 
human movements~\cite{HoldenSK16, autocompletion18}. 
More recently, there has been some work focusing on how to achieve a smoother transition~\cite{RobustTweening20}.
These studies mainly focus on predicting and controlling cyclic actions such as walking and running, 
and they often achieve better results in those simple movements. 
However, when applied to complex and impromptu activities such as dance performance or martial arts, 
the generated animations are far from satisfactory.

Compared with walking, running, and other cyclic locomotion, learning to synthesize diverse and artistically-elegant 
dance movements from keyframes is more challenging.
Firstly, dance performance is highly irregular with complex kinetics, \emph{e.g.}, 
body rotation in contemporary dance can have various speeds, different strengths, and larger amplitudes. 
Secondly, dance movements are inherently diversified, \emph{e.g.}, 
the current motion can be followed by a wide range of possible movements.
Thirdly, in long-term motion sequence synthesis, the entire dance movements may be composed of different dance 
action units or various combinations of them.
Lastly, it is more difficult to obtain a dance performance than walking or running. 
A well-choreographed dance animation requires collaborative efforts between 
animators, dancers, and choreographers, which is an expensive and tedious process. 
However, little research investigates efficient keyframe-based motion synthesis for an impromptu dance performance at present. 
The topic is extremely valuable, which can significantly reduce the demand for professional motion capture systems, 
the dependence on professional choreographers, and the workload of animation designers when creating new dance animation. 

In this paper, we propose a novel keyframe-based motion generation network based on multiple constraints, 
which can achieve diverse dance synthesis via learned knowledge. 
The constraints include the given keyframes, the root trajectory, and the velocity factor.
Similar to~\cite{TimeWarping20}, the velocity factor is designed to constrain the motion synthesis.
Specifically, the approach is mainly formulated based on the recurrent neural network (RNN) and the Transformer architecture.
The core of our network is a hierarchical RNN module composed of two long short-term memory (LSTM) units, 
in which the first LSTM is utilized to embed the posture information of the historical frames into a latent space, 
and the second one is employed to predict the human posture for the next frame. 
We also design two Transformer-based controllers to model the constraints of the root trajectory 
and the velocity factor respectively. 
The self-attention layers in the Transformer encourage a model considering the broad context in a given sequence 
by learning the relationships between different elements~\cite{Transformer17}.
Therefore, the proposed Transformer-based controllers enable our network to utilize the temporal 
context of the frames to achieve fine-grained motion control.
We verify the proposed algorithm on a dance dataset containing a variety of contemporary dance. 
The video and quantitative analyses prove the superiority of our algorithm. 
Moreover, the demo and qualitative experimental results demonstrate that the complex motion 
sequence generated by our algorithm is capable of producing diverse and smooth motion transitions between keyframes, 
even for long-term synthesis. 

In summary, \textbf{our main contributions} are listed as follows:
\begin{itemize}
\item
We propose a novel neural network based on LSTM and Transformer for complex motion generation via keyframes. 
The model is elaborately controlled under the root trajectory and the velocity factor constraints,
and can generate complex dance movements satisfying the control conditions. 
\item
We design the velocity factor constraint for fine-grained dance motion synthesis. 
By specifying the velocities of different body parts, 
our model is able to enhance the diversity and smoothness 
in long-term motion generation.
\item
Compared with the state-of-the-art motion transition methods, the data synthesized by our technique 
on the dance dataset obtain better accuracy in terms of various evaluation criteria, 
and the quality of character animation is also higher.
\end{itemize}

\section{Related work}
In this section, we briefly review the literatures closely related to our work, 
including human motion modeling, dance motion synthesis, and motion transition generation. 

\subsection{Human motion modeling based on deep learning}
Recently, deep learning has gained remarkable success in the field of both CV and CG. 
To synthesis realistic and natural human motion,
there has been a surge in modeling human motion with neural networks.
For example, Holden \emph{et al.}~\cite{HoldenSK16} propose a deep convolutional network to
perform human motion synthesis and resolve the ambiguity via incorporating foot contact information.
However, the proposed framework concentrates on simple character movements regarding walking, running, and punching.
%
Inspired by the image inpainting techniques, Hernandez \emph{et al.}~\cite{STinpaint19} 
reformulate motion prediction as an inpainting task to complete the masked joints in spatiotemporal volumes. 
It is well-known that recurrent neural networks (RNNs) have inherent advantages in modeling sequential data. 
Hence recent work employs RNNs to model human motion.
Martinez \emph{et al.}~\cite{MartinezB017} make several modifications to the standard RNN models and develop a
 sequence-to-sequence architecture with residual connections for short-term human motion prediction. 
Lee \emph{et al.}~\cite{LeeLL18} present a multi-layer RNN conditioned to handle spatiotemporal
constraints and structural variabilities for interactive character animation. 
Wang \emph{et al.}~\cite{STML21} formulate a new spatiotemporal RNN framework to investigate the motion manifold. 
The proposed network avoids the generation of average posture and eliminates the need for a separate disambiguous network.
Also, Wang \emph{et al.}~\cite{RNN-AT21} construct an RNN-based generator for human motion synthesis and utilize a 
refiner network with adversarial training loss to refine motion sequences. 
In our work, the RNN is utilized for sequence modeling and long-term motion synthesis. 
The constrained conditions in our method is modeled using two additional Transformers, 
which enables the network to synthesize diverse dance motion sequence.

\subsection{Dance motion synthesis}
Dance motion synthesis can be viewed as a typical conditional motion generation task.
Due to the high correlation between dance and music, extensive works have been dedicated to 
the music-oriented dance generation. 
In the early research, dance-to-music is regarded as the problem of template matching, which attempts to
generate dance movements according to musical similarity~\cite{DanceAnimation06, MusicSim13}.
By analyzing the rhythmic patterns of motions, Kim \emph{et al.}~\cite{motionBeat} facilitate 
the rhythmic motion generation synchronized with an input sound signal.
However, the above methods are limited by their capacity of the provided dataset. 
Recently, neural networks dominate dance motion synthesis.
Tang \emph{et al.}~\cite{lstmMelody18} design an LSTM-autoencoder model to extract the mappings between
music and motion, which can largely enhance choreography in accordance with the music. 
Lee \emph{et al.}~\cite{danceToMusic19} and Ye \emph{et al.}~\cite{ChoreoNet20} 
attempt to synthesize dance from music 
through a two-stage procedure, which firstly learn basic dance units and then organizes the basic units into dance sequence.
Nevertheless, two-stage generation methods lack enough flexibility and scalability.
To alleviate error accumulation of autoregressive model in long-term motion generation, 
Huang \emph{et al.}~\cite{huang2021} formulate the music-conditioned dance generation as a
sequence-to-sequence learning problem and utilize the curriculum learning strategy to enhance the training process.
Considering that the motion manifolds of classical convolutional and recursive neural models are 
non-Euclidean geometry, Ferreira \emph{et al.}~\cite{GCNdance21} design a novel method based on graph convolutional 
networks to synthesize human motion from music.
%
%
The goal of our research is different from the above work. We attempt to synthesize dance motion sequences 
conditioned on temporal sparse keyframes with user-specified root trajectory and velocity factor. It is a
rarely explored but rather valuable topic for its significance in character 
animation and entertainment.

\subsection{Motion transition generation}
In computer animation, there have been intensive investigations on motion transition generation.
We limit the task as synthesizing intermediate movements between user-specified keyframes.
It is quite challenging as significant motion gaps must be filled under sparse temporal constraints.
Pioneering approaches are mainly based on retrieval paradigms, which search the matched motion clips 
from a database and blend them, such as motion graphs~\cite{MotionGraph02, MotionGraph09}. 
After that, methods based on probabilistic models have been widely adopted for human motion.
Chai \emph{et al.}~\cite{MAP07} and Min \emph{et al.}~\cite{MAP09} formulate motion synthesis as 
maximum a posterior (MAP) problem.
Wang \emph{et al.}~\cite{GPDM08} apply Gaussian process dynamical models (GPDMs) 
for learning models of human pose and motion. 
However, the above methods are designed for designated actions, making the algorithms 
look like pre-arranged scripts and rules that will fail in confronting complex human movements.
Because of the impressive scalability and expressiveness of deep neural networks, recent studies tend to
 use deep learning for motion transition prediction. 
Gaisbauer \emph{et al.}~\cite{NPB19} present a fully-connected feed-forward neural network for 
generating feasible postures from given input postures.
Zhang \emph{et al.}~\cite{autocompletion18} formulate an autoregressive recurrent neural network (ARNN) 
that is conditioned on the target keyframes for motion-aware animation with fixed interpolated frames.
Harvey \emph{et al.}~\cite{RAT18} propose the recurrent transition networks (RAT) based on LSTM to synthesize 
missing data between keyframes. However, the method can still only generate motion transition sequences 
with definite lengths and is limited to periodic routine actions.
Later, Harvey \emph{et al.}~\cite{RobustTweening20} improve the RAT method~\cite{RAT18} by adding the 
time-to-arrival embedding to the network, which allows the method generates variable transition lengths. 
They verify the method on a periodic dance dataset and obtain good animation results.
In summary, existing methods mainly focus on the motion transition of periodic actions with relatively short-term synthesis,
while our work concentrates on the long-term and impromptu contemporary dance generation. 
For diverse motion synthesis, besides the conventional root trajectory constraint, 
we introduce the velocity factor for the first time in motion modeling, 
which preserves the naturalness and diversity of the generated dance movements.

\section{Method}
Throughout the paper, we denote $keyframes$ as the user-specified temporal sparse representative frames in a motion sequence. 
Our generative model can be regarded as a time series modeling and synthesis problem. 
The overall framework of dance synthesis is illustrated in Figure~\ref{fig:demo}.
With the current frame and the constrained conditions, the model can automatically predict 
the pose and the associated information of the next frame. 
Given the keyframes and the length of the sequence to be predicted,
the system can automatically synthesize the intermediate motion sequence between the keyframes. 
Moreover, users can specify the trajectory of the root joint and the moving speeds of different body parts, 
which are employed as control signals to guide the network to generate the desired animation. 

\begin{figure*}[htb]
\centering
\includegraphics[width=1.0\linewidth,height=0.43\linewidth]{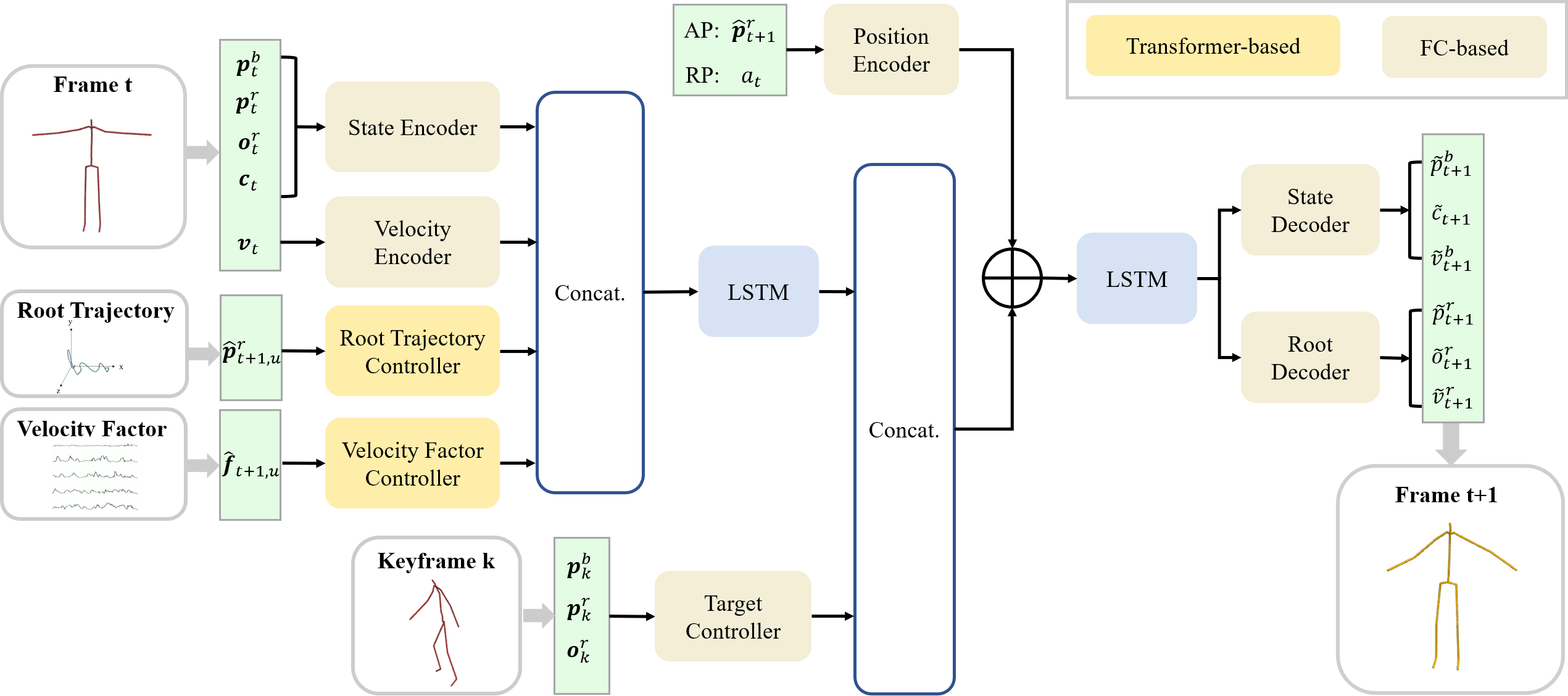}
   \caption{\label{fig:framework}
     The framework of our motion synthesis network. It shows all the computations for a single timestep.}
\vspace{-3mm}
\end{figure*}
\subsection{Variable definitions}
The dance dataset we use comes from\cite{dataset1, dataset2}. 
We remove the finger joints, and the resulted human skeleton contains $23$ joints. 
We preprocess the dataset by rotating the postures at different timesteps toward the positive direction 
of $z$-axis and extract the related information based on the rotated postures. 
The definition of the symbols used in this paper is listed in Table~\ref{tab:SymDef}.

\begin{table}[!htbp]
\centering
\caption{A table with variable definitions.}
\begin{tabularx}{0.495\textwidth}{lX}
\hline
\textbf{Symbol} &  \textbf{Definition} \\
\hline
 $t$ & The time mark of the current frame.\\
 $J$ & The total number of joints. \\
 $k_1$, $k_2$ & The indexes of the given keyframes, where $k_1 < k_2$.\\
  $N$ & Number of frames between keyframes, $N=k_2 - k_1$.\\
 $\textbf{o}^r_t$& The rotation angle to rotate a posture toward the positive direction of the $z$ axis.\\
 $\textbf{p}^r_t$& The global position of the root joint.\\
 $\textbf{p}^b_t$& The root-relative position of other joints except for root joint. The dimension is $(J-1)\times 3$.\\
 $\textbf{p}_t$  & The set of the global position of root joint and the root-relative positions of other joints, 
 that is $\textbf{p}_t=\{\textbf{p}^r_t, \textbf{p}^b_t\}$.\\
 $\textbf{v}^r_t$& The velocity of the root joint.\\
 $\textbf{v}^b_t$& The velocities of other joints except for root joint. The dimension is $(J-1)\times 3$.\\
 $\textbf{v}_t$  & The set of velocities of all the joints with $\textbf{v}_t=\{\textbf{v}^r_t, \textbf{v}^b_t\}$.\\
 $\textbf{c}_t$  & The foot contact labels of two toe joints and two heel joints with dimension of 4.\\
 $\textbf{f}_t$  & The velocity factor vector with dimension of 5.\\
 $\hat{\textbf{p}}^r$& The given root trajectory sequence with dimension of $N\times 3$.\\
 $\hat{\textbf{f}}$  & The given velocity factor sequence with dimension of $N\times 5$.\\
 $\hat{\textbf{p}}^r_{t,u}$ & A subsequence of the given root trajectory sequence. The subsequence 
 is centered at frame $t$, the window size is $u$, and the dimension is $u\times 3$. \\
 $\hat{\textbf{f}}_{t,u}$& A subsequence of the given velocity factor sequence. The subsequence 
 is centered at frame $t$, the window size is $u$, and the dimension is $u\times 5$.\\
 $M$ & The number of divided body parts.\\
\hline
\end{tabularx}%
\label{tab:SymDef}
\end{table}%

We use $\textbf{X}_t=\{\textbf{p}_t, \textbf{o}^r_t, \textbf{c}_t, \textbf{v}_t\}$ to represent the 
 ground truth information at time $t$ and $\tilde{\textbf{X}}_t=\{\tilde{\textbf{p}}_t, \tilde{\textbf{o}}^r_t, 
 \tilde{\textbf{c}}_t, \tilde{\textbf{v}}_t\}$ to denote the prediction results of the network at time $t$. 
 The given inputs and conditions can be expressed as $\{ \textbf{p}_{k_1}, \textbf{o}^r_{k_1}, \textbf{p}_{k_2}, 
 \textbf{o}^r_{k_2}, \hat{\textbf{p}}^r, \hat{\textbf{f}} \}$. The output of the model contains 
 the position sequence $\{\tilde{\textbf{p}}_{k_1+1}, \cdots,\tilde{\textbf{p}}_{k_2} \}$ of all joints,
 the rotation sequence $\{\tilde{\textbf{o}}^r_{k_1+1}, \cdots, \tilde{\textbf{o}}^r_{k_2} \}$ of the root joint, 
 the contact label sequence $\{\tilde{\textbf{c}}_{k_1+1}, \cdots, \tilde{\textbf{c}}_{k_2} \}$ of the toe joints and heel joints, 
 as well as the velocity sequence $\{\tilde{\textbf{v}}_{k_1+1}, \cdots, \tilde{\textbf{v}}_{k_2} \}$ of all joints.

\subsection{The proposed motion synthesis network}
The proposed network aims to receive historical information and control signals, 
then produce smooth and diverse dance sequences. 
As shown in Figure ~\ref{fig:framework}, we demonstrate the generation process in one timestep.
Given two keyframes $k_1$ and $k_2$, the problem is
how to generate a realistic motion sequence between the two keyframes with a natural transition. 
In essence, this is a time-series-prediction problem. 
Inspired by~\cite{RobustTweening20}, we utilize the LSTM unit to predict the motion sequence.

{\flushleft\textbf{Overview of the network.}}
The whole framework consists of three encoders, three controllers, and two decoders. 
The three encoders are comprised of a state encoder, a velocity encoder, and a position encoder. 
Specifically, the state encoder receives the posture information, 
\emph{i.e.}, the root-relative positions $\textbf{p}^b_t$,
 the root's global position $\textbf{p}^r_t$, the root's rotation angle $\textbf{o}^r_t$, 
 and the foot contact information $\textbf{c}_t$. 
The position encoder takes the next frame's root position as input to make the network aware of the temporal embedding.
The velocity encoder projects the dynamics of the movement to our network.
The three controllers are composed of a root trajectory controller, a velocity factor controller, 
and a target controller.
The root trajectory controller takes the positions of the root's trajectory as input 
to ensure the generated motion moves towards the specified trajectory.
The velocity factor is proposed to achieve fine-grained motion control by specifying the moving speeds of different body parts. 
The target controller receives the posture information of the keyframe. 
The goal is to make the network perceive the distance between the predicted frame and the target keyframe.
The decoders contain a root decoder and a state decoder. 
The root decoder is in charge of decoding the root position, rotation, and velocity information, 
while the state decoder predicts the relative positions and velocities of other joints except for the root, 
as well as the foot contact states at time $t+1$. In summary, the three encoders, two decoders, and the target controller are constructed based on 
the fully connected layer (FC-based) with different depths, while the root trajectory controller and the velocity factor controller 
are formulated based on the Transformer structure (Transformer-based).

\begin{figure}[htbp]
\centering
\includegraphics[width=0.67\linewidth,height=0.75\linewidth]{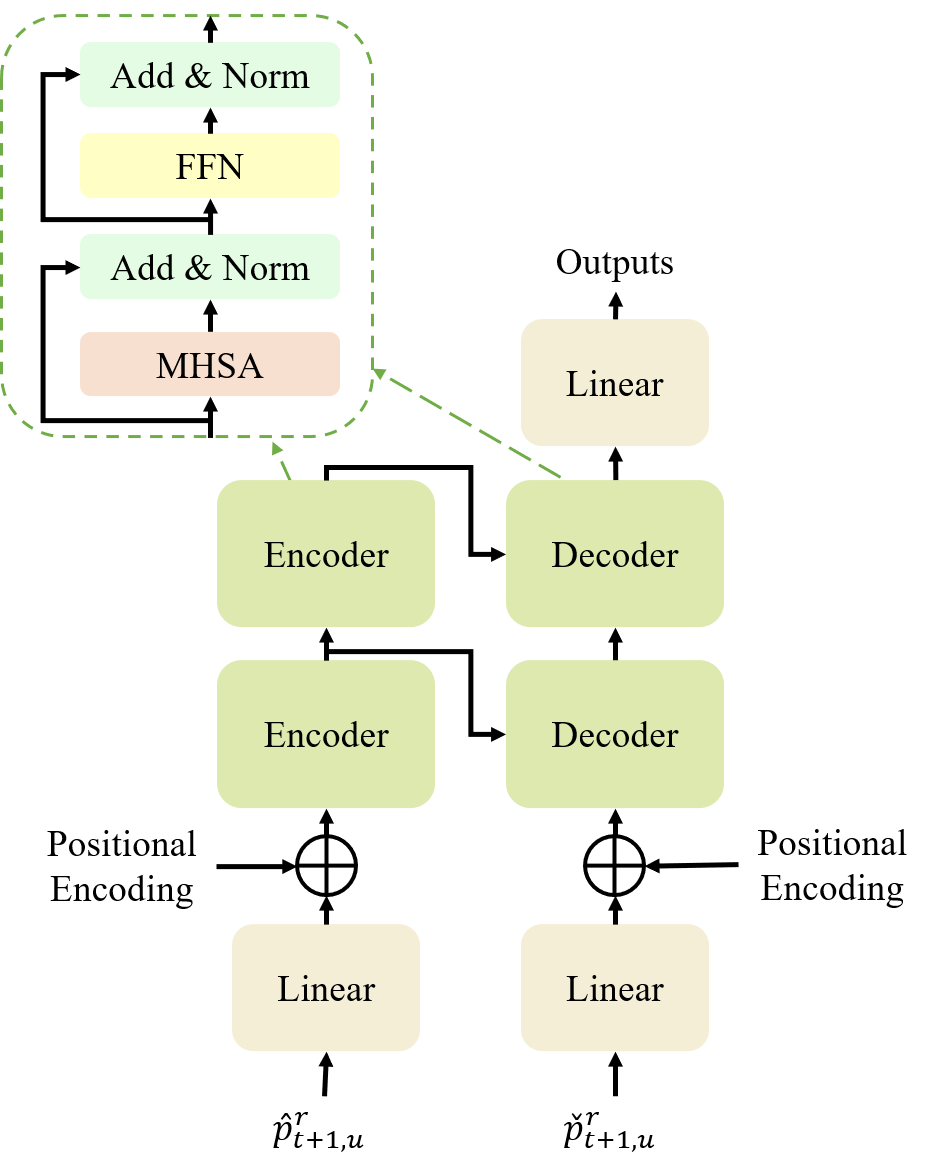}
   \caption{\label{fig:Trans}
     The structure diagram of the Transformer-based root trajectory controller.}
\vspace{-3mm}
\end{figure}

{\flushleft\textbf{The root trajectory controller.}}
To make the generated motion consistent with the given trajectory, it is crucial to let 
the network capture the temporal context information from the past to the future.
Since Transformer~\cite{Transformer17} can well model long-range dependency by leveraging 
the query-key correlation to different tokens, it has shown outstanding results in 
natural language processing~\cite{NLP19}, neural machine translation~\cite{NMT20}, 
and various vision tasks~\cite{FPT20,transOD20}. 
Therefore, we construct the root trajectory controller based on the Transformer mechanism.

We illustrate the detailed structure of the root trajectory controller in Figure~\ref{fig:Trans}, where  
$\hat{\textbf{p}}^r_{t+1,u}$ refers to the target root trajectory segment with frame $t+1$ as the center and the window size of $u$. 
$\check{\textbf{p}}^r_{t+1,u}$ is the modification of $\hat{\textbf{p}}^r_{t+1,u}$, that is, 
the data before the frame $t+1$ is replaced by the predicted root trajectory. 
Considering that the root trajectory predicted by the network cannot be completely consistent with as the target trajectory, 
and the current predicted position is not only required to smoothly connect with the past predicted results, 
but also move along the future target trajectory, it is not enough to  only encode the given target trajectory.
Therefore, we introduce the $\check{\textbf{p}}^r_{t+1,u}$ and define it as the mixture root trajectory segment. 
The encoders and decoders in the Transformer are the standard combinations of the multi-head self-attention (MHSA) 
and the feed forward networks (FFN). The MHSA comprises multiple self-attention blocks 
and explicitly models the interactions between all entities of a sequence. 
The FFN is used to perform information transformation. 
The residual add and layer norm are implemented after both the MHSA and the FFN.
 
For the constraint coding process, the controller first accepts $\hat{\textbf{p}}^r_{t+1,u}$ as input, 
projects it into the embedding space, 
and then extracts the trajectory representation with two encoders. We also project 
$\check{\textbf{p}}^r_{t+1,u}$ into the feature space and then decode it with two decoders. 
Skip connections are used between the encoders and the decoders to increase the flexibility of information flow. 
As pointed out in~\cite{Transformer17}, the Transformer has a permutation invariance. We follow its 
procedure, incorporating the positional encoding into $\hat{\textbf{p}}^r_{t+1,u}$ and $\check{\textbf{p}}^r_{t+1,u}$ 
to make the controller aware of the trajectory segments' positions. 
The technical details can be found in the literature above.
Finally, the decoded result is transformed by a linear layer to acquire the representation
 of the root trajectory constraint. 

\begin{figure}[htbp]
\centering
\begin{tabular}{c@{ }c}
\includegraphics[width=0.45\linewidth, height=.46\linewidth]{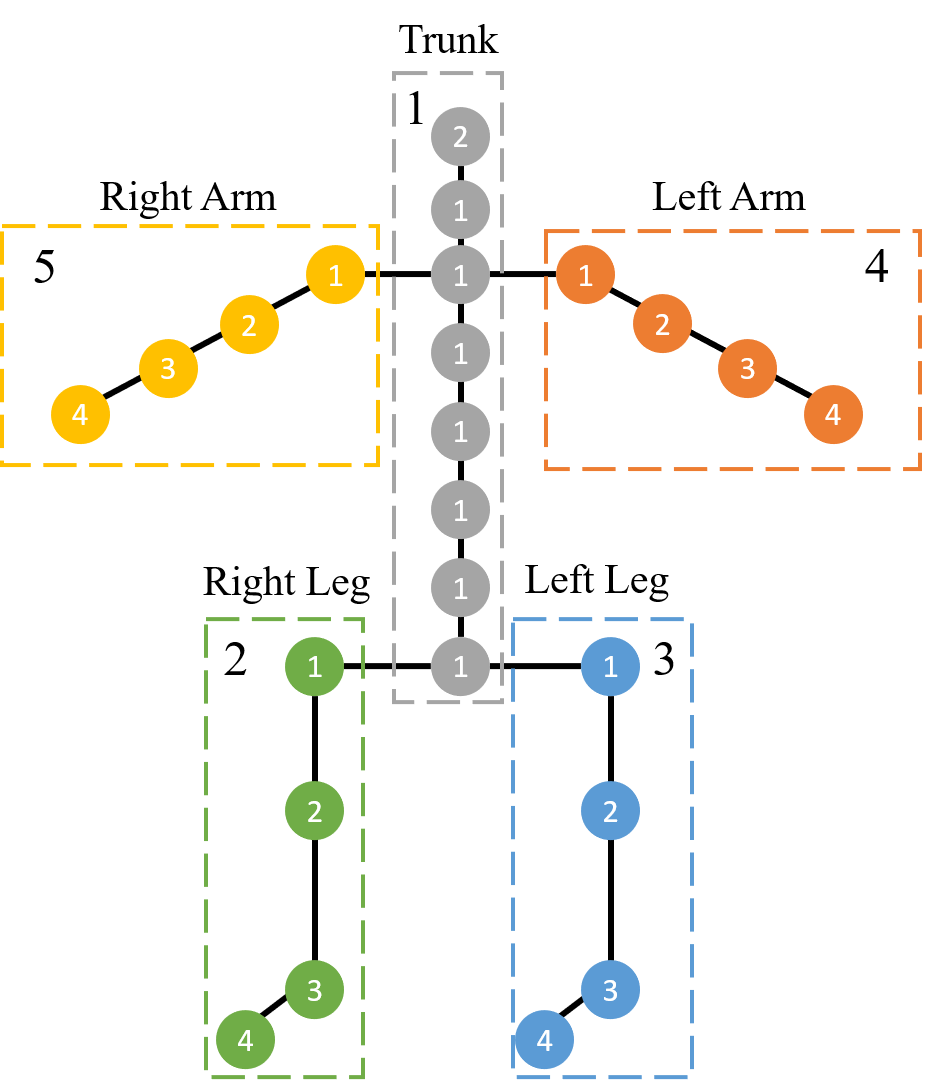}&
\includegraphics[width=0.55\linewidth, height=.4\linewidth]{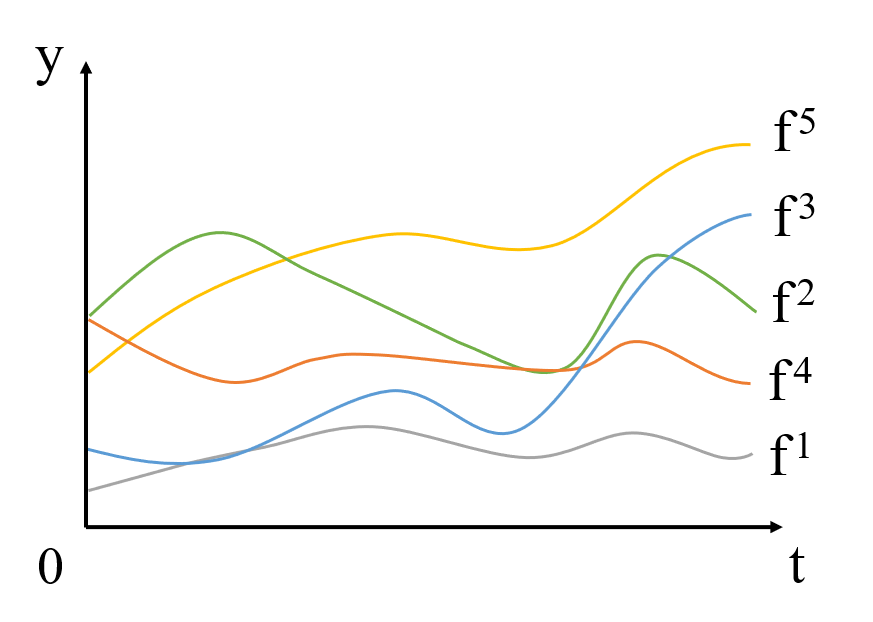}\\
(a) Body part division & (b) Velocity factor sequences\\
\end{tabular}
\caption{\label{fig:body}%
        The structure diagram representation of the velocity factor. (a) Different colors represent different body parts. 
     The numbers in the circles represent the joint weights, while the numbers in the dotted boxes denote the body-part index. 
     (b) The values of the corresponding velocity factors are displayed in this figure.}
\vspace{-1mm}
\end{figure}

{\flushleft\textbf{The velocity factor controller.}}
To achieve fine-grained control, the velocity factor controller is introduced.
As the speeds of different body parts can vary significantly when dancing, 
we divide the human body into five parts: the trunk, the left and right arms, 
the left and right legs. 
We get five velocity factors by weighted average over the moving speeds of the joints related to the body parts.
Figure~\ref{fig:body} (a) displays the structure diagram of body division, 
where different colors represent different body parts.
For different joints, the weights are set according to their distances from the end effector.
The numbers in the circle of Figure~\ref{fig:body} (a) are the weights used in this paper.
In general, the end effectors are more influential in describing the dynamics of a movement, 
so joints closer to them have larger weights.  
The velocity factor of body part $i$, $i\in \{1, 2, 3, 4, 5\}$ at frame $t$ can be calculated as:
\begin{equation}
\label{eq1}
\setlength{\abovedisplayskip}{3pt}
\setlength{\belowdisplayskip}{3pt}
f^i_t= \frac{\sum_j^{J_i}(w^{ji} \times \|\textbf{v}_t^{ji}\|_2)}{\sum_j^{J_i}w^{ji}},
\end{equation}
where $J_i$ is the joint set of part $i$, $w^{ji}$ is the weight of joint $j$ in part $i$, 
and $\textbf{v}_t^{ji}$ is the velocity of joint $j$ in part $i$ at time $t$. 
The obtained velocity factor sequences of different body parts for a given motion sequence are 
illustrated in figure~\ref{fig:body} (b).

Considering human motion at consecutive moments should be continuous without sudden changes, 
it is important for the velocity factor controller to also perceive the long-range context information, 
so as to learn the smooth temporal context representation. 
Therefore, we construct the velocity factor controller with the same structure as the trajectory controller. 
The differences are that the inputs $\hat{\textbf{p}}^r_{t+1,u}$ and $\check{\textbf{p}}^r_{t+1,u}$ are replaced 
by the $\hat{\textbf{f}}_{t+1,u}$ and $\check{\textbf{f}}_{t+1,u}$. 
Here, the $\hat{\textbf{f}}_{t+1,u}$ is a sequence segment of the given velocity factor with the 
time $t+1$ as the center and the window size $u$, and the $\check{\textbf{f}}_{t+1,u}$ replaces the data of $\hat{\textbf{f}}_{t+1,u}$ 
before time $t+1$ with the network predicted results. 
Similar to $\check{\textbf{p}}^r_{t+1,u}$, $\check{\textbf{f}}_{t+1,u}$ is defined as the mixture velocity factor segment.

{\flushleft\textbf{The LSTM backbone.}}
If we directly send all the coding formation of the encoders and the controllers to the LSTM network, 
the information received by the network will be too much and too complex. 
However, one LSTM unit may not be able to process so much data at one time. 
Therefore, we employ two LSTM units hierarchically to receive and process the encoded information. 
The first LSTM receives the outputs of the state and velocity encoders, as well as the root trajectory and  
velocity factor controllers.
After the information is processed by the first LSTM, we concatenate its output with the keyframe embedding 
of the target controller and send the concatenated results to the second LSTM for further processing. 
The sequential processing inputs and control conditions make the network achieve better results. 
We will validate it in the ablation study.
In experiments, we employ a scheduled sampling mechanism to select the input of LSTM in each timestep. 
Specifically, we first define a sampling probability rate $\beta$. 
When predicting actions in different timesteps, we select the ground truth as the input with the probability of $\beta$. 
At the beginning, $\beta$ is set as 1. With the progress of the training, $\beta$ decays in an exponential decay manner. 
The learning strategy can make the convergence of the network more stable and make the motion synthesis 
and prediction more smooth.

When the pose information of the target keyframe is directly sent to the network, 
the network cannot know the temporal distance between the frame to be predicted and the target keyframe. 
Referring to the insight in Transformer, we encode the position information of the frame to be predicted
and add it with the input of the second LSTM unit. We use one fully connected layer to embed the position information.
Specifically, as pointed out in~\cite{PointTrans}, the 3D joint coordinate is a kind of natural position coding. 
We define the root's position as the absolute position (AP) representation. 
The root joint of frame $t+1$ extracted from the root trajectory constraint is used as input of the position encoder. 
Since the trajectory sequence of the root joint may have the same position at different times, we design 
an additional relative position (RP) representation $a_t$, which is calculated as follows:
\begin{equation}
\label{eq:pos}
\setlength{\abovedisplayskip}{3pt}
\setlength{\belowdisplayskip}{3pt}
a_t= \frac{t+1-k_1}{k_2-k_1}.
\end{equation} 

At last, the state decoder and the root decoder are used to predict the posture information of the next movement. 
However, when all the information is decoded by the state decoder, the predicted position sequence of the root joint 
is discontinuous with relatively large fluctuation. 
Inspired by the 3D human body pose estimation algorithms~\cite{pose3D1,pose3D2}, 
we separate the information of the root joint and use the root decoder to predict its state information. 
This practice can make the generated trajectory sequence smoother.

\subsection{Loss functions}
According to the network structure and goal, we define several loss functions, including the 
reconstruction loss $\mathcal{L}_{rec}$, posture consistency loss $\mathcal{L}_{con}$, 
root trajectory smooth loss $\mathcal{L}_{root}$, keyframe consistency loss $\mathcal{L}_{key}$, 
and velocity factor consistency loss $\mathcal{L}_{vfac}$.  
The complete loss function is defined as follows:
\begin{equation}
\label{eq:loss}
\setlength{\abovedisplayskip}{3pt}
\setlength{\belowdisplayskip}{3pt}
\mathcal{L}\!=\! w_{rec}\mathcal{L}_{rec}\!+\!w_{con}\mathcal{L}_{con}\!+\!w_{root}\mathcal{L}_{root}
\!+\!w_{key}\mathcal{L}_{key}\!+\!w_{vfac}\mathcal{L}_{vfac},
\end{equation}
where $w_{rec}$, $w_{con}$, $w_{root}$, $w_{key}$, $w_{vfac}$ are the corresponding loss weights. 
We set the loss weights as 0.3, 0.2, 0.15, 0.2, and 0.15 respectively in training phase. 
These parameters are obtained through a number of experiments.
We give the details of each loss below.

{\flushleft\textbf{Reconstruction loss.}}
The mean square error (MSE) loss is taken to construct the reconstruction loss, 
which can force the network to generate motion sequences satisfying the designed control constraints. 
The reconstruction loss $\mathcal{L}_{rec}$ can be expressed as: 
\begin{equation}
\label{eq:rec}
\setlength{\abovedisplayskip}{3pt}
\setlength{\belowdisplayskip}{3pt}
\mathcal{L}_{rec}= \frac{1}{N}\sum^{k_2}_{t=k_1+1}\|\tilde{\textbf{X}}_t-\textbf{X}_t\|^2,
\end{equation}
where $N$ is the sequence length. The terms of $\mathcal{L}_{rec}$ include the reconstruction of joint positions, 
root joint rotation angles, foot contact labels, and joint velocities at each moment. 

{\flushleft\textbf{Posture consistency loss.}}
When the network is trained only with individual joints, the correlations between the connected joints 
are neglected to a certain extent. Hence, we introduce the bone length consistency loss $\mathcal{L}_{bone}$ to force 
the network to generate posture consistent with the ground truth bone length. 
In the meantime, the foot contact labels and the joint velocities can be inferred from the predicted 
posture information. Thus, we also introduce the foot contact consistency loss $\mathcal{L}_{contact}$ 
and the joint velocity consistency loss $\mathcal{L}_{velocity}$ 
to avoid the contradiction of prediction information. 
Therefore, the consistency loss $\mathcal{L}_{con}$ consists of $\mathcal{L}_{bone}$, $\mathcal{L}_{contact}$ 
and $\mathcal{L}_{velocity}$, and they are expressed as follows: 
\begin{align}
\label{eq:con}
\setlength{\abovedisplayskip}{3pt}
\setlength{\belowdisplayskip}{3pt}
&\mathcal{L}_{con}= \mathcal{L}_{bone}+\mathcal{L}_{contact}+\mathcal{L}_{velocity},\\
&\mathcal{L}_{bone}= \frac{1}{N}\sum^{k_2}_{t=k_1+1}\big(\sum_{(i,j)\in \mathcal{B}}\big\| \|\tilde{\textbf{p}}_t^i-\tilde{\textbf{p}}_t^j\|_2-l_{ij}\big\|^2\big),\\
&\mathcal{L}_{contact}= \frac{1}{N}\sum^{k_2}_{t=k_1+1}\big(\sum_i^{\mathcal{F}}\tilde{c}_t^i \|\tilde{\textbf{v}}_t^i\|_2\big),\\
&\mathcal{L}_{velocity}= \frac{1}{N}\sum^{k_2}_{t=k_1+1}\big(\sum_i^{\mathcal{J}}\|\tilde{\textbf{v}}_t^i-(\tilde{\textbf{p}}_t^i-\tilde{\textbf{p}}_{t-1}^i)\|^2\big),
\end{align}
where $\mathcal{B}$ in $\mathcal{L}_{bone}$ is the index set consisted of all natural connected 
joint pairs in the human skeleton. 
$l_{ij}$ is the original length of the bone segment formed by joint $i$ and $j$, the length of which 
can also be obtained by calculating the distance between $\tilde{\textbf{p}}_t^i$ and $\tilde{\textbf{p}}_t^j$. 
We penalize the length inconsistency between the ground truth and the inferred to force 
the correctness of the predicted posture information.
In $\mathcal{L}_{contact}$, $\mathcal{F}$ is the index set of the foot contact joints, 
$\tilde{c}_t^i$ is the predicted contact label for foot joint $i$ at frame $t$, 
$\tilde{c}_t^i=1$ if there is foot contact and $0$ otherwise.
We impose a penalty on the product of the L2-norm of $\tilde{\textbf{v}}_t^i$ and the corresponding contact label 
$\tilde{c}_t^i$ to force the consistency of the predicted foot contact label and the predicted velocity~\cite{STML21}.
In $\mathcal{L}_{velocity}$, $\mathcal{J}$ is the index set of all joints. 
The velocities of different joints at frame $t$ can be inferred by subtracting skeleton positions in the previous frame 
from the current frame $t$. 
The information consistency can be effectively guaranteed by punishing the differences between the inferred velocities 
and the predicted velocities of our network.

{\flushleft\textbf{Root trajectory smooth loss.}}
We refer to the long horizon loss function in~\cite{STML21} and extract the loss term of root joint to form
our root trajectory smooth loss $\mathcal{L}_{root}$:
\begin{equation}
\label{eq:root}
\setlength{\abovedisplayskip}{3pt}
\setlength{\belowdisplayskip}{3pt}
\mathcal{L}_{root}= \frac{1}{N} \left(\sum^{k_2}_{t\!=\!k_1\!+\!1}\|\tilde{\textbf{p}}^r_t\!-\!\tilde{\textbf{p}}^r_{t-1}\|^2 \!+\!
\sum^{k_2}_{t\!=\!k_1\!+\!1}\|\tilde{\textbf{o}}^r_t\!-\tilde{\textbf{o}}^r_{t-1}\|^2 \right),
\end{equation}
We minimize the differences between the root's spatial positions and rotation angles 
at frame t and t -1 to enforce temporal consistency. When we impose a similar constraint on other joints, 
 the network tends to overfit, and the final result converges to an average posture.  
 Therefore, we only use the joint trajectory smooth loss to constrain the natural smooth transition for the root joint.
 
{\flushleft\textbf{Keyframe consistency loss.}}
One of the main tasks of the network is to build natural connections of the generated dance 
with the given keyframes, which means that we need to ensure the continuity of the predicted movements 
near keyframes and at keyframes. Hence, the keyframe consistency loss $\mathcal{L}_{key}$ is introduced to achieve the goal:
\begin{small}
\begin{equation}
\label{eq:key}
\mathcal{L}_{key}\!=\!\left\{
\begin{aligned}
& \frac{1}{2m} \! \left( \! \sum^{k_1\!+\!m}_{t\!=\!k_1\!+\!1} \|\tilde{\textbf{p}}_t\!-\!\textbf{p}_{k_1}\|^2 \!+\! \sum^{k_2}_{t\!=\!k_2\!-\!m\!+\!1} \|\tilde{\textbf{p}}_t\!-\!\textbf{p}_{k_2}\|^2 \!\right), & \!N\!>\!2m, \\
& \! \frac{1}{N}\!\sum^{k_2}_{t\!=\!k_1\!+\!1} \!\left( \frac{t\!-\!k_1}{N} \| \!\tilde{\textbf{p}}_t\!-\!\textbf{p}_{k_1}\|^2 \!+\! 
\big(\!1\!-\!\frac{t\!-\!k_1}{N}\big) \| \!\tilde{\textbf{p}}_t\!-\!\textbf{p}_{k_2}\|^2 \!\right), & \!N \! \leq \! 2m,
\end{aligned}
\right.
\end{equation}
\end{small}
where $m$ is the number of frames affected by the keyframes. 
When $N>2m$, we impose constraints on the $m$ frames near the keyframe $k_1$ or keyframe $k_2$; 
When $N \leq 2m$, the in-between $N$ frames are constrained to be affected by the mixture results 
of the keyframes $k_1$ and $k_2$. 
We use $\frac{t-k_1}{N}$ as the impact factor to determine the influence weights of two keyframes on the 
predicted pose at time $t$.
However, when we only use the information of keyframes to calculate the keyframe consistency loss, 
there is a significant discontinuity at keyframes. 
By also imposing a temporal consistency constraint on the postures near the keyframes, 
the generated dance sequence can achieve smooth transitions at the given keyframes. In the training process, we set $m=5$. 
In experiments, we will validate the superiority of the proposed loss function.

{\flushleft\textbf{Velocity factor consistency loss.}}
In order to make the velocity factors of the synthesized dance sequences consistent with the given control condition, 
the velocity factor consistency loss $\mathcal{L}_{vfac}$ is proposed. After obtaining the network's outputs, 
we calculate the velocity factor $\tilde{\textbf{f}}_t$ for each frame of the generated dance motion. 
We formulate the velocity factor consistency loss as follows:
\begin{equation}
\label{eq:vfac}
\setlength{\abovedisplayskip}{3pt}
\setlength{\belowdisplayskip}{3pt}
\mathcal{L}_{vfac}= \frac{1}{N}\sum^{k_2}_{t=k_1+1}\|\tilde{\textbf{f}}_t-\hat{\textbf{f}}_t\|^2, 
\end{equation}
where $\hat{\textbf{f}}_t$ is the constrained velocity factor given by the user.

\subsection{Traning details}
The dance database we use contains a total of 123 pieces of contemporary dance.
Among them, $80\%$ of the data set, a total of 98 dance segments, 93347 frames are used for training, 
and the remaining $20\%$ of data, a total of 25 segments, 20897 frame samples are used for testing. 
For the network, the state encoder, the velocity encoder, 
and the target controller are all composed of two fully connected layers. 
The hidden units for the two layers are set as 512 and 256, respectively. 
The position encoder consists of one fully connected layer with 512 hidden units. 
For the Transformer-based controllers, \emph{i.e.}, the root trajectory controller and the velocity factor controller, 
the temporal window size $u$ is 7, the head number and the embedding dimension for 
the MHSA block are 8 and 32 respectively, and the dimensions for the three linear layers are set as 256. 
For the two LSTM blocks, the hidden units are set to 256. 
Both the state decoder and the root decoder consist of three fully connected layers, where 
the dimensions for the two hidden layers are 512 and 256, respectively. 
We use the Parametric Rectified Linear Unit (PReLU)~\cite{PReLU15} as the activation function 
for all the encoders, decoders, and the target controller, 
while for the two Transformer-based controllers, we follow the same activation function and 
structure as the typical Transformer does~\cite{Transformer17}.

We implement our model using PyTorch, and all models used in this paper are trained on a GeForce RTX 2080 Ti GPU. 
When training the model, we use the Adam optimizer~\cite{adam} and set the learning rate to 0.0001 and the batch size to 128. 
It takes 43 hours to train the proposed network and 0.0085 seconds to generate a single frame during testing.
Since our model can generate variable-length motion transition sequences, we also use variable-length 
dance sequences to train the network. 
The minimum sequence length $min\_len$ used in training is $5$ and the maximum sequence length $max\_len$ is $70$. 
For each epoch, we set the current $min\_len$ and $max\_len$ to represent the sequence length range of the current epoch. 
At the early stages of training, both $min\_len$ and $max\_len$ are set to $5$, and the $max\_len$ increases by 1 after each epoch.
In this way, the amount of training data in each round gradually increases as the training progresses.
To prevent repeated learning on dance sequences of small lengths, we let $min\_len$ increase by $4$ every five epochs. 
When $max\_len$ is greater than $70$, the training process is terminated. 
\section{Experiments and results}
\subsection{Different control constraints}
Our model can achieve fine-grained control for dance movement generation through the skillfully designed control constraints. 
We conduct experiments to verify the effects of different control constraints with only one of the constraints 
changed at a time.

\begin{figure}[htb]
\centering
\vspace{-2mm}
\includegraphics[width=1.\linewidth, height=.35\linewidth]{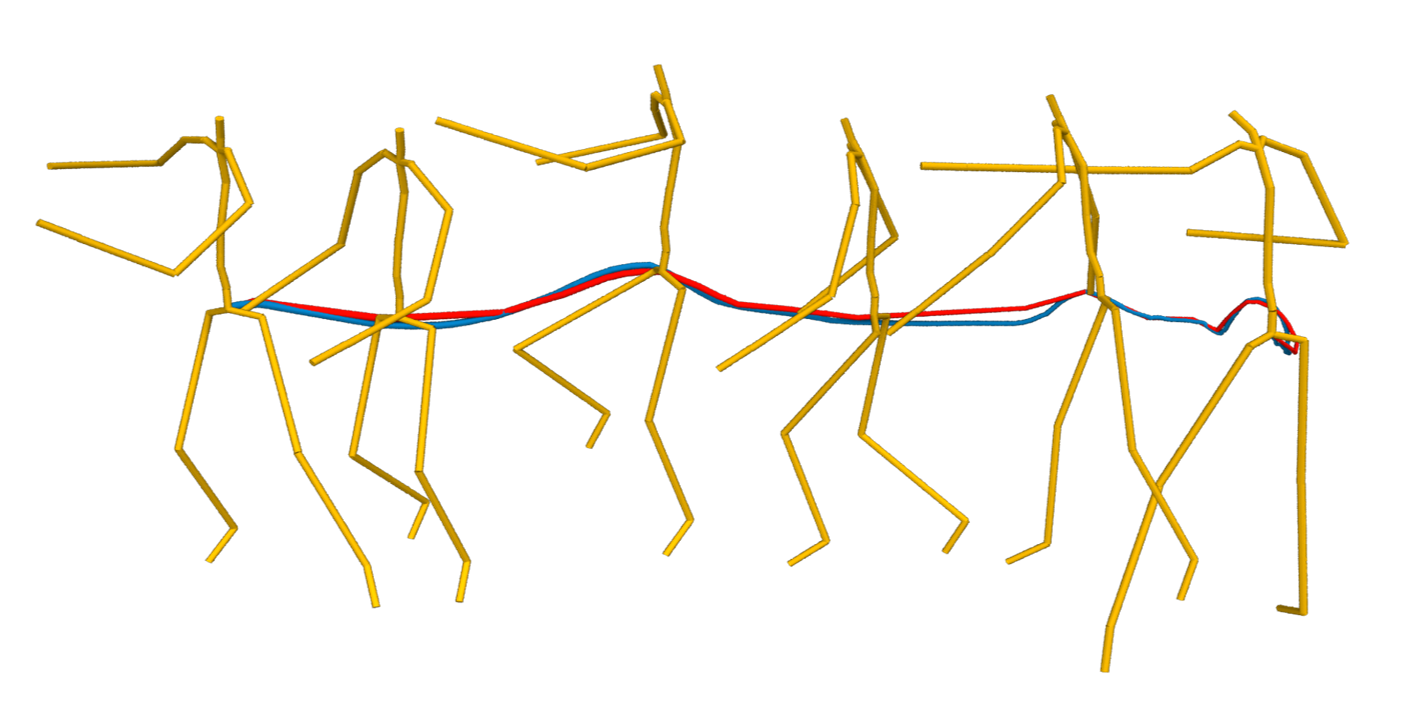} \\
(a) Motion sequence\\
\includegraphics[width=.7\linewidth, height=.22\linewidth]{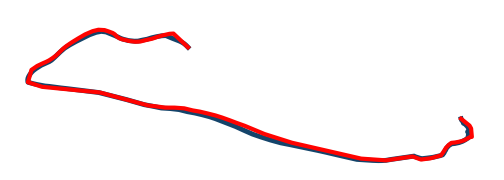}\\
(b) Root trajectory\\
\caption{\label{fig:root_traj}%
          Visualization of the generated motion sequence (a) and the related root trajectories (b), 
          where the target trajectory is blue and the generated trajectory is red.}
\vspace{-3mm}
\end{figure}

{\flushleft\textbf{Different root trajectories.}}
In order to verify the effect of the root trajectory, we visualize the generated root trajectory and 
the specified control trajectory to compare the difference between them in Figure~\ref{fig:root_traj}. 
It can be observed that the motion sequence generated by our model can well fit the root trajectory 
conditions specified by the user.
Meanwhile, we also defined a quantitative index to evaluate the accuracy of the predicted root trajectory. 
The quantitative index is explained in the ablation experiment (Section~\ref{sec:abalation}). 
\begin{figure}[htb]
\centering
\begin{tabular}{c@{ }c}
\includegraphics[width=0.47\linewidth, height=.28\linewidth]{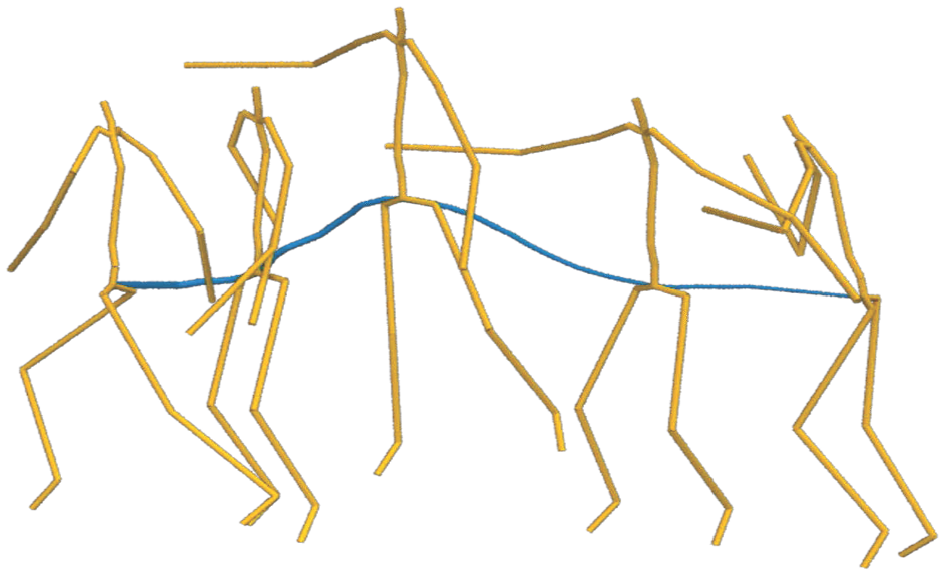}&
\includegraphics[width=0.47\linewidth, height=.26\linewidth]{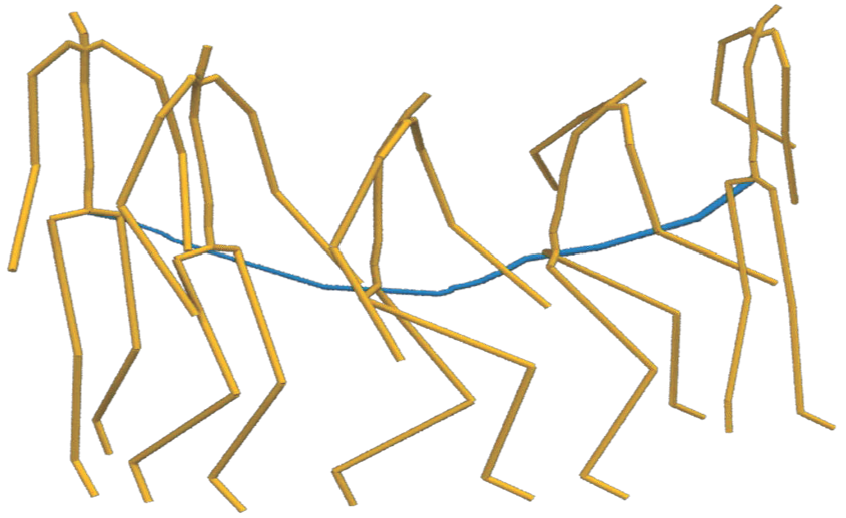}\\
(a) Jump & (b) Squat\\
\end{tabular}
\caption{\label{fig:root_traj2}%
           Visualization of the generated motion sequences under different root trajectories.}
\vspace{-3mm}
\end{figure}
\begin{figure}[htb]
\centering
\includegraphics[width=1.0\linewidth, height=.2\linewidth]{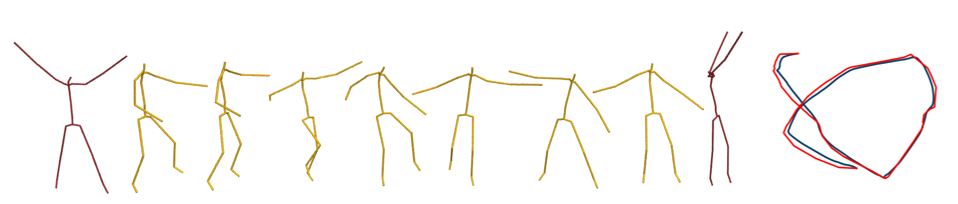}\\
(a) Root trajectory 1\\
\includegraphics[width=1.0\linewidth, height=.2\linewidth]{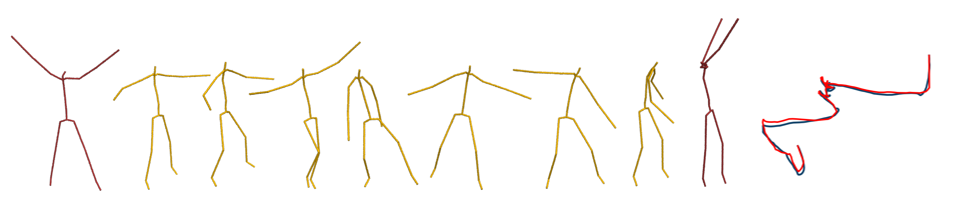}\\
(b) Root trajectory 2 \\
\caption{\label{fig:root_traj3}%
           Visualization of the generated motion sequences (the first and last postures are the keyframe postures) 
           under different root trajectories (the red one is the generated motion and the blue one is the ground truth) 
           with other conditions the same.}
\vspace{-3mm}
\end{figure}

When the user provides different root trajectories or different root heights, it should produce different actions.
For example, if the root joint reaches a high position, a jump action should be generated, 
and a squat action should be produced if the root joint is low.
The phenomenons can be observed in our results (Figure~\ref{fig:root_traj2}), which demonstrates 
the close correlations between the generated motions and the height of the root joint. 
Furthermore, we conduct experiments with totally different root trajectories while maintaining the same keyframes 
and the velocity factors. 
The results are illustrated in Figure~\ref{fig:root_traj3}. 
We can observe that various transition movements are generated, and the diversity of dance motion 
is significantly enhanced under the control of the root trajectory.
The above experiments validate that the root trajectory constraint can well control the global position of joints, 
affect the types of actions, and promote the diversity of dances.

\begin{figure*}[htbp]
\centering
\includegraphics[width=0.95\linewidth, height=.45\linewidth]{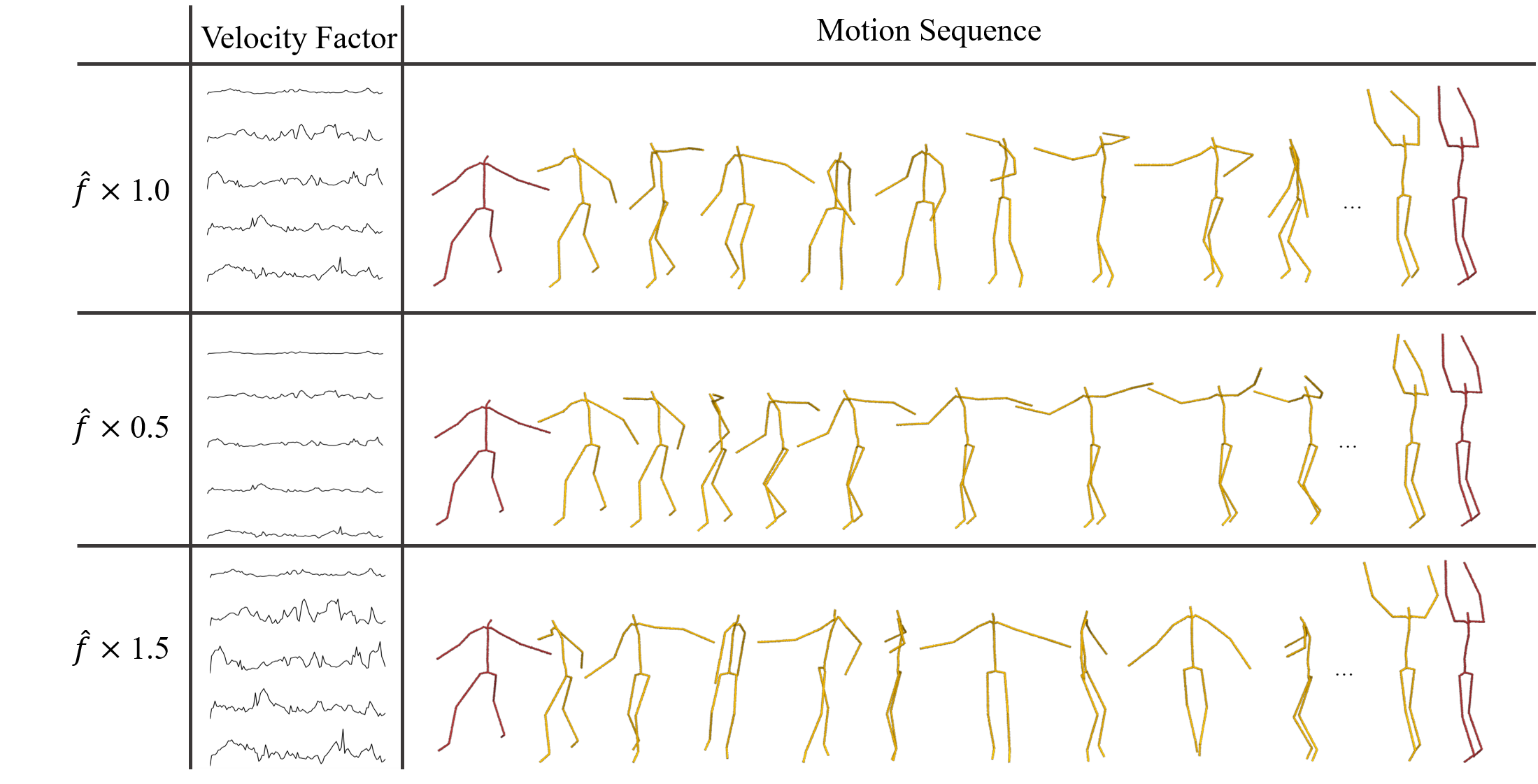} 
\caption{\label{fig:Vecfac}%
        Visualization of the generated motion sequences under different velocity factors with other conditions the same, where 
         the first and last postures are the keyframe postures.}
\vspace{-3mm}
\end{figure*}

{\flushleft\textbf{Different velocity factors.}}
When other conditions are the same, that is, the keyframes are the same, and the root trajectory 
constraint is also the same, we change the values of velocity factors to generate different motion sequences. 
When using different velocity factor sequences, motion sequences with different speeds will be generated. 
In order to enable users to control the overall speed better, we allow users to use a constant 
value to achieve multiple changes to the velocity factor sequences. 
Figure~\ref{fig:Vecfac} illustrates the generated dance sequences with the velocity factors varied according to 
the multiples of 1.0, 0.5, and 1.5. We visualize the generated motion movements every five frames. 
From Figure~\ref{fig:Vecfac} and the supplementary video, we observe that when the velocity factors become 
half of the original, the overall speed of the motion sequence becomes very slow, 
and the variety of actions is significantly reduced. 
Besides, the motion movements tend to go directly from one keyframe to another keyframe.
However, when the velocity factors increase 1.5 times, the overall movements become faster and change dramatically. 
Because of the enlarged velocity factors, the generated action will not slide directly from one keyframe to another keyframe, 
and the diversities of intermediate movements have been promoted to a large extent.  
 
\begin{figure}[htbp]
\centering
\includegraphics[width=1.0\linewidth, height=.2\linewidth]{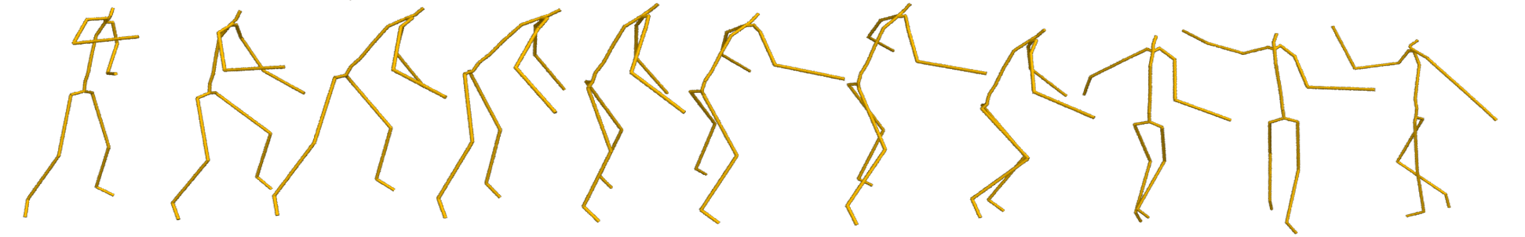}\\
(a) Original velocity factors\\
\includegraphics[width=1.0\linewidth, height=.2\linewidth]{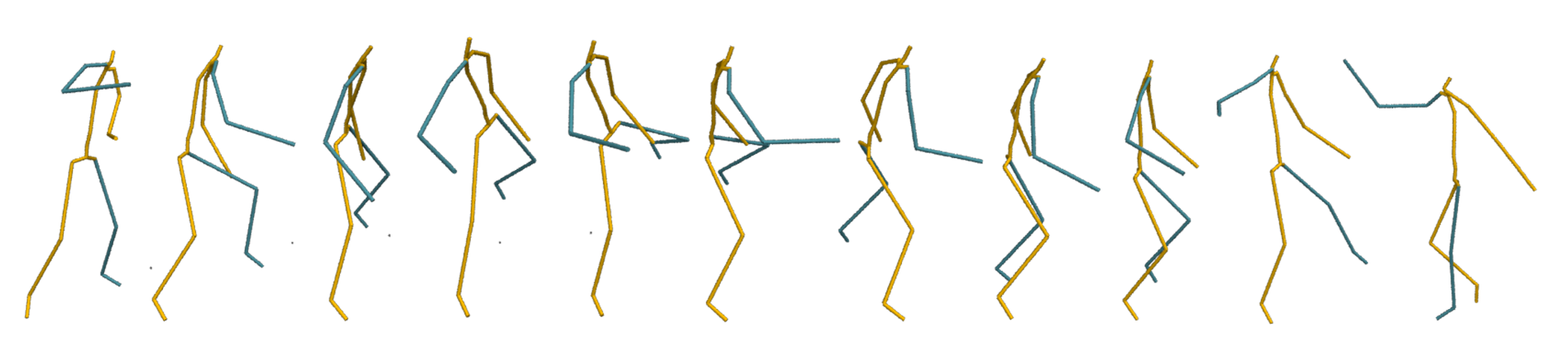}\\
(b) 1.5 $\times$ velocity factors (blue parts)  \\
\caption{\label{fig:Vecfac2}%
           Visualization of the generated motion sequences under different velocity factors 
           with other conditions the same.}
\vspace{-3mm}
\end{figure}

The above experimental results illustrate that the same multiple changes are applied to the five velocity factors. 
Beyond that, users can assign different velocity factors to different body parts to achieve more fine-grained motion control. 
We implement experiments to change the velocity factors of different body parts. 
The generated dance movements are displayed every five frames in Figure~\ref{fig:Vecfac2}, 
where Figure~\ref{fig:Vecfac2} (a) is the synthesized motion sequence 
with the original velocity factors, while Figure~\ref{fig:Vecfac2} (b) is the generated dances 
with the blue body parts (right arm and left leg) higher velocity factors. 
It can be observed that the blue parts generate more active results compared with the yellow parts.
The experiments further validate that the diversities of synthesized dances can be enhanced 
by varying the velocity factor constraint.

\begin{figure}[htbp]
\centering
\includegraphics[width=1.0\linewidth, height=.6\linewidth]{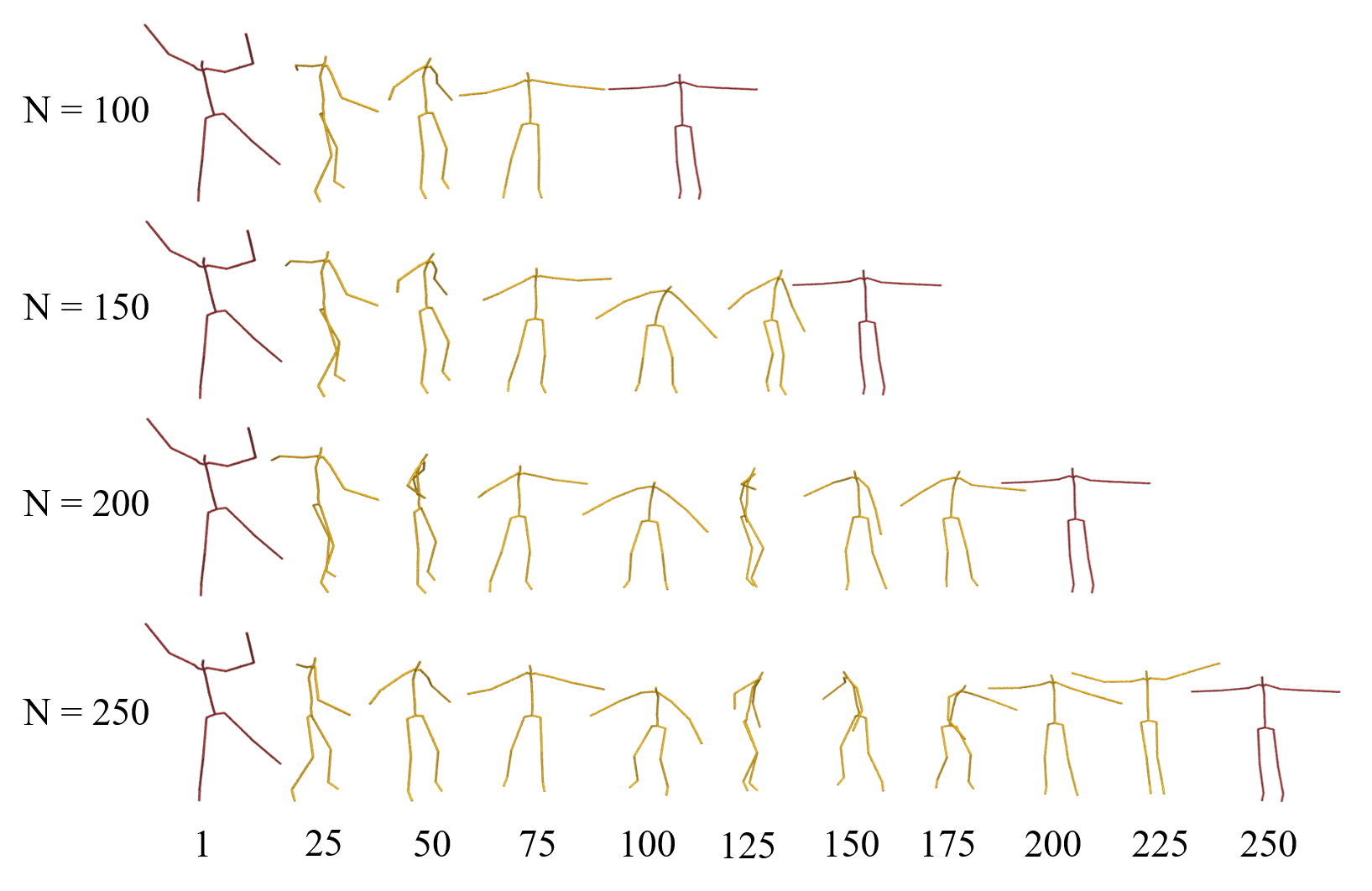}\\
\caption{\label{fig:DiffLen}%
          Visualization of the generated motion sequences (the first and last postures are the keyframe postures) 
           under different lengths with other conditions the same.}
\vspace{-5mm}
\end{figure}

{\flushleft\textbf{Motion transitions with different lengths.}}
To verify the generation ability of our model and the diversities of generated motions, we use the same keyframes 
to generate long-term motion transition sequences of 100, 150, 200, and 250 frames, respectively. 
We visualize the four generated dance sequences in 25 frames in Figure~\ref{fig:DiffLen}.
It turns out that the intermediate actions are still diversified when the length of the generated sequence is 200 frames. 
When the length exceeds 200 frames, repeated meaningless actions begin to appear, which can be seen in the last row of 
Figure~\ref{fig:DiffLen} and the supplementary video. 
The results further reflect the importance of the two control conditions, \emph{i.e.}, 
the root trajectory and the velocity factor, to the generation of the action. 
The existence of the two terms can make the synthesized action more diversified and realistic.

\subsection{Ablation study}
\label{sec:abalation}
We conduct ablation experiments to verify the superiority of different modules of our network.
We define three quantitative metrics: the evaluation criterion of joint position, 
the evaluation criterion of velocity factor, and the evaluation criterion of root trajectory. 
The units of the three metrics are cm, cm/frame and cm, respectively.

Inspired by the L2P criteria~\cite{RobustTweening20}, we propose the average L2 distances of 
root-relative position (LRP) to measure the deviation between the predicted motions and their ground truth:
\begin{equation}
\label{eq:lrp}
\setlength{\abovedisplayskip}{3pt}
\setlength{\belowdisplayskip}{3pt}
LRP = \frac{1}{\mathcal{|D|}} \frac{1}{N} \sum_{c\in\mathcal{D}} \sum_{t=k_1+1}^{k_2} \|\tilde{\textbf{p}}_t^b(c)-\textbf{p}_t^b(c)\|_2, 
\end{equation}
where $\mathcal{D}$ is a test set, $\mathcal{|D|}$ is the number of $\mathcal{D}$, and 
$c$ is a transition sequence of $\mathcal{D}$. 
$\tilde{\textbf{p}}_t^b(c)$ refers to the root-relative positions of skeleton joints at time $t$ in sequence $c$, 
and $\textbf{p}_t^b(c)$ is the related ground truth.
The smaller the distance error is, the more accurate the predicted movements are. 

We use the accuracy rates for the evaluation indicators for both velocity factor and root trajectory.
We define two fault-tolerant thresholds, $\delta_v$ (in cm/frame) and $\delta_r$ (in cm), to calculate the corresponding accuracies.  
We calculate the difference between the ground truth and the inferred result from the predicted 
motion sequence for the velocity factor. 
If the difference is less than $\delta_v$, the result is correct; otherwise, it is considered to
be out of the range of correct values. 
The average accuracy of the velocity factor (AVF) can be obtained by:
\begin{gather}
\label{eq:avf}
AVF=\frac{1}{\mathcal{|D|}} \sum_{c\in\mathcal{D}}
\begin{cases}
	1, & if \quad g_v(c)<\delta_v,\\
	0, & otherwise,
\end{cases}\\
g_v(c) = \frac{1}{N}\frac{1}{M} \sum_{t=k_1+1}^{k_2}\sum_{i=1}^{M} |\tilde{f}_t^i(c)-\hat{f}_t^i(c)|, 
\end{gather}
where $g_v(c)$ is the deviation of the velocity factor for dance sequence $c$, 
$\tilde{f}_t^i(c)$ is the predicted value of body part $i$ at timestep $t$ in sequence $c$, 
and $\hat{f}_t^i(c)$ is the corresponding ground truth.

The average accuracy of the root trajectory (ART) can be obtained similarly as the AVF:
\begin{align}
\label{eq:art}
&ART = \frac{1}{\mathcal{|D|}} \sum_{c\in\mathcal{D}}
\begin{cases}
 1, & if \quad g_r(c)<\delta_r, \\
 0, & otherwise,
\end{cases}\\
&g_r(c) = \frac{1}{N} \sum_{t=k_1+1}^{k_2} \|\tilde{\textbf{p}}_t^r(c)-\hat{\textbf{p}}_t^r(c)\|_2, 
\end{align}
where $g_r(c)$ indicates the deviation of root trajectory for dance sequence $c$, 
$\tilde{\textbf{p}}_t^r(c)$ is the predicted root joint position at timestep $t$ of sequence $c$, 
and $\hat{\textbf{p}}_t^r(c)$ is the ground truth.
During the experiment, we take $\delta_v$ as 1.0 and $\delta_r$ as 7.0. 

We report the ablation results of the three quantitative metrics under different model settings 
in Table~\ref{tab:pos}, Table~\ref{tab:vel}, and Table~\ref{tab:traj}. 
Specifically, the ``One LSTM'' model represents that the generation model contains one LSTM unit to 
deal with postural information and control conditions.
``Condition-FC'' refers to replace the Transformer-based controllers with two fully connected layers to encode the 
constraints of root trajectory and velocity factor. 
``One decoder'' means that the state decoder decodes the posture information of all skeleton joints, 
including the root joint.
``Without Velfac constraint'' stands for the model without the velocity factor constraint.
``Without $\mathcal{L}_{key}$'' denotes we train the whole model without the keyframe consistency loss $\mathcal{L}_{key}$.
``Whole model'' is the overall framework learned with all loss functions.
``Interpolation'' is the model that utilizes the interpolation strategy
to synthesize transition action between keyframes.
At last, Harvey's method~\cite{RobustTweening20} in terms of LRP in Table~\ref{tab:pos} has also been reported.

\begin{table}[htb]
\caption{The LRP evaluation for the transition sequence at the length of 10, 50, 100, 150, 
and the average result (AVG). The best results are shown in bold.}
\vspace{-3mm}
\begin{center}
\resizebox{0.49\textwidth}{!}
{
\begin{tabular}{|c|c|c|c|c|c|}
\hline
\multirow{2}{*}{Models} & \multicolumn{4}{c|}{Frames} & \multirow{2}{*}{AVG}\\
\cline{2-5}
  & 10 & 50 & 100 & 150  &\\
\hline
Interpolation   &  \textbf{13.83}  &  87.59  &  113.59  &124.47	&84.87     \\
\hline
Harvey \emph{et al.}~\cite{RobustTweening20} &  141.79	&116.40	&201.43	&297.22	&189.21   \\
\hline
One LSTM & 32.81 & 70.49  & 88.11 & 100.81 & 73.06  \\
\hline
Condition-FC   & 37.30	&67.03	&85.20	&96.39	&71.48    \\
\hline
One decoder   & 28.36	&51.15	&65.04	&77.24	&55.45   \\
\hline
Without Velfac constraint   &  34.37	&62.30	&83.81	&97.52	&69.50\\
\hline
Without $\mathcal{L}_{key}$   &    41.59&	82.95	&97.31	&105.46	&81.83 \\
\hline
Whole model  &  27.07&	\textbf{50.06}&	\textbf{63.01}&	\textbf{74.37}&	\textbf{53.63}  \\
\hline
\end{tabular}
}
\end{center}
\label{tab:pos}
\vspace{3mm}
\caption{The 
AVF evaluation for transition sequence at the length of 10, 50, 100, 150, 
and the average result (AVG). The best results are shown in bold.}
\vspace{-3mm}
\begin{center}
\resizebox{0.49\textwidth}{!}
{
\begin{tabular}{|c|c|c|c|c|c|}
\hline
\multirow{2}{*}{Models} & \multicolumn{4}{c|}{Frames} & \multirow{2}{*}{AVG}\\
\cline{2-5}
  & 10 & 50 & 100 & 150  &\\
\hline
One LSTM &   0.68&	0.75&	0.70&	0.67&	0.70  \\
\hline
Condition-FC   &   0.68	&0.71	&0.68	&0.65	&0.68  \\
\hline
One decoder   & 0.71	&0.75	&0.73	&0.70	&0.72   \\
\hline
Without Velfac constraint    & 0.60	&0.60	&0.53	&0.49	&0.56  \\
\hline
Without $\mathcal{L}_{key}$   &   0.66	&0.68	&0.66	&0.64	&0.66  \\
\hline
Whole model  &   \textbf{0.72}	&\textbf{0.76}	&\textbf{0.75}	&\textbf{0.72}	&\textbf{0.74}  \\
\hline
\end{tabular}
}
\end{center}
\label{tab:vel}
\vspace{3mm}
\caption{The 
ART evaluation for transition sequence at the length of 10, 50, 100, 15, 
and the average result (AVG). The best results are shown in bold.}
\vspace{-3mm}
\begin{center}
\resizebox{0.49\textwidth}{!}
{
\begin{tabular}{|c|c|c|c|c|c|}
\hline
\multirow{2}{*}{Models} & \multicolumn{4}{c|}{Frames} & \multirow{2}{*}{AVG}\\
\cline{2-5}
  & 10 & 50 & 100 & 150  &\\
\hline
One LSTM &  0.63& 	0.74& 	0.76	& \textbf{0.74}& 	0.72   \\
\hline
Condition-FC   &  0.47&	0.64	&0.67	&0.67	&0.61   \\
\hline
One decoder   & 0.63	&0.55	&0.48	&0.39	&0.51   \\
\hline
Without Velfac constraint    &  0.62	&0.59	&0.45	&0.34	&0.50  \\
\hline
Without $\mathcal{L}_{key}$   &   0.62	&0.42	&0.33	&0.27	&0.41 \\
\hline
Whole model  &    \textbf{0.86}	& \textbf{0.85}	& \textbf{0.80}	& 0.72	& \textbf{0.81} \\
\hline
\end{tabular}
}
\end{center}
\label{tab:traj}
\end{table}

{\flushleft\textbf{Effectiveness of two LSTM units.}}
To verify the effectiveness of using two LSTM, we conduct experiments by learning the model with one LSTM. 
It turns out that the dance sequence generated with one LSTM performs less well than the ``Whole model'', 
leading to much larger LRP errors. The results can be observed in Table~\ref{tab:pos}. 
For the velocity factor, the accuracy produced by the ``Whole model'' is much higher than the ``One LSTM'', 
as shown in Table~\ref{tab:vel}.
As for the root trajectory, when the sequence length is less than 100, the accuracy of the ``Whole model'' 
far outstrips the model learned with one LSTM. However, when the predicted length starts larger than 150, 
the ``Whole model'' is slightly worse than ``One LSTM''. 
The reason may be that one LSTM unit cannot reach a balance between various control conditions. 
Instead, it tends to dominate the control of the root trajectory. Hence the results of posture and velocity factor 
become worse. On the contrary, by processing different information hierarchically and sequentially, 
the whole model with two LSTM reaches a good balance between different constraints 
and achieves better results on the three criteria.

{\flushleft\textbf{Effectiveness of Transformer controller.}}
Due to the multi-head self-attention layer mechanism in Transformer, the Transformer-based controllers 
can well leverage the context information in the time window of the frame to be predicted to encode the control signals.
We use the fully connected layer to replace the Transformer to validate its importance on constraint control.
By comparing the ``Condition-FC'' with the ``Whole model'' in Table~\ref{tab:pos}, Table~\ref{tab:vel}, and 
Table~\ref{tab:traj}, we observe that the performance of the complete model is far better than that of 
the model constraint with the fully connected layer in terms of the joint position error, 
velocity factor and root joint prediction accuracy, demonstrating the superiority of Transformer 
structure in constraints modulation.

\begin{figure}[htb]
\centering
\includegraphics[width=1.\linewidth, height=.18\linewidth]{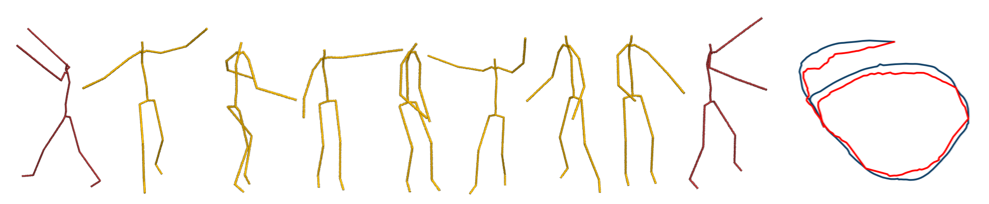} \\
(a) One decoder\\
\includegraphics[width=1.\linewidth, height=.18\linewidth]{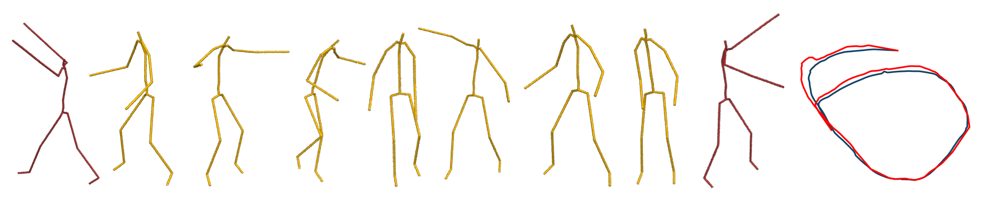}\\
(b) Two decoders\\
\caption{\label{fig:decoder}%
           Visualization of the generated motion sequences (the first and last postures are the keyframe postures) 
           and root trajectories (the red one is the generated motion and the blue one is the ground truth).}
\vspace{-3mm}
\end{figure}

{\flushleft\textbf{Effectiveness of the root decoder.}}
Existing methods usually employ the state decoder to predict the posture information of 
all skeleton joints~\cite{RobustTweening20, RAT18}. 
However, for dance, the action space of dance movements is vast. Only one decoder may cause large 
jitter in the generated root trajectory. 
To verify the effectiveness of the root decoder, we conduct experiments to 
 obtain the prediction results with only the state decoder. 
 Comparing the experimental results of ``One decoder'' and ``Whole model'' in Table~\ref{tab:pos}, Table~\ref{tab:vel}, and 
Table~\ref{tab:traj}, we can observe that the  LRP and the AVF of ``One decoder'' are 
similar to those of the complete model, but the 
ART is inferior. 
We also illustrate the synthesized dance sequences and the corresponding root trajectories 
of ``One decoder'' and ``Whole model'' in Figure~\ref{fig:decoder}.
It can be observed that the root trajectory obtained by using one decoder has a larger deviation 
from the target root trajectory, while the result obtained by using two decoders is more
consistent with the target root trajectory, which validates the advantages of the root decoder.

{\flushleft\textbf{Effectiveness of velocity factor constraint.}}
We implement experiments to verify the affect of the velocity factor constraint in motion synthesis
by removing the corresponding controller. It can be seen from Table~\ref{tab:vel} when velocity factors 
are not used for control, the AVF evaluation metric is $56\%$. 
After adding the velocity factor constraint, the AVF performance 
has improved to $74\%$. 
In addition, the whole model has significantly improved the model ``Without Velfac constraint'' 
in terms of the evaluation criteria of LRP and AVT 
as illustrated in Table~\ref{tab:pos} and Table~\ref{tab:traj}. 
Three quantitative experimental results confirm the importance of the velocity factor constraint.

{\flushleft\textbf{Effectiveness of keyframe consistency loss.}}
Keyframe consistency loss aims to constrain the motion sequence transited natural and smooth near keyframes.
To verify its importance, we conduct experiments by learning the network without the $\mathcal{L}_{key}$ function. 
In the experiment, we find that the performance of the three quantitative evaluation indexes of the model 
is greatly improved by adding the keyframe consistency loss. 
The detailed results are illustrated in Table~\ref{tab:pos}, Table~\ref{tab:vel}, and Table~\ref{tab:traj}. 
The achieved significant gains demonstrate that the keyframe consistency loss has a positive effect on the 
optimization of the network.

\subsection{Comparison with other methods}
{\flushleft\textbf{Experiments on dance dataset.}}
We also compare our model with two related methods to validate its superiority. 
Specifically, we make a comparison with Harvey's method~\cite{RobustTweening20} and the results 
obtained by interpolation strategy.
When training Harvey's model~\cite{RobustTweening20}, we employ the same learning strategy on the dance dataset we use.  
It turns out that the generated motion sequences are far from satisfactory, many synthesized actions are unnatural,
 and the continuity at the keyframes is poor. 
 The main reason for the results may be that the network is mainly designed for walking, running, and other simple movements, 
 while the dance movements are relatively complex. 
 Its variability makes it impossible to uniquely describe the movements using only keyframes, 
 which may easily cause ambiguity. Therefore, the final synthesized results are performed poorly. 
For the interpolation strategy, we take the root trajectory and keyframes as the control conditions. 
We spherically interpolate the quaternions between the keyframes to complete the motion transition.
 
We also calculated the LRP errors of the two methods. 
The comparison results are reported in Table~\ref{tab:pos}. 
It can be observed from Table~\ref{tab:pos} that within short-term motion transition, \emph{e.g.} $N\leq10$, 
the interpolation method shows a decisive advantage. 
It is partly because the motion becomes almost linearly in a sufficiently short timescale. 
However, with the increasing length of the prediction sequence, our method obviously exceeds the interpolation-based strategy. 
In addition, the results in Table~\ref{tab:pos} also prove that our method is always better than 
Harvey's model~\cite{RobustTweening20} under different prediction sequence lengths and far exceeds it 
by a large margin. 

\begin{figure}[htbp]
\centering
\includegraphics[width=1.0\linewidth, height=.6\linewidth]{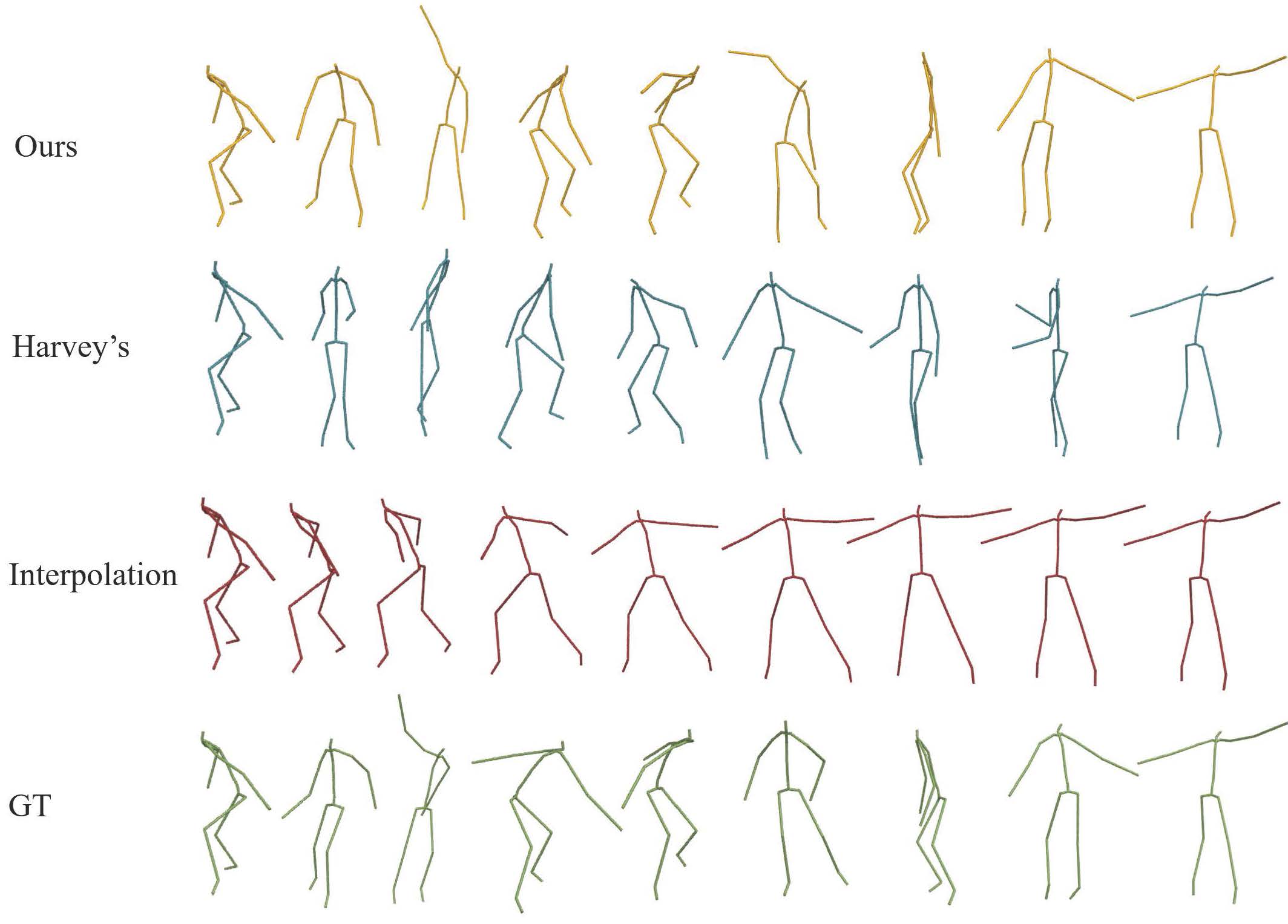}\\
\caption{\label{fig:Compare}%
           Visualization of the generated dance sequences by different methods and the ground truth.}
\vspace{-3mm}
\end{figure}

We show the visual comparison of the dance sequences generated by our algorithm and 
the above two algorithms, as well as the ground truth (GT) sequences in Figure~\ref{fig:Compare}. 
It can be seen that the motion generated by Harvey's method~\cite{RobustTweening20} are noisy and 
unnatural, and there are obvious discontinuities at the keyframes. 
The results obtained by interpolation lose the diversity of actions, and the steps are always floating. 
In contrast, our method improves the problem above and can generate more continuous and smooth motion movements.
The qualitative and quantitative comparative experiments demonstrate that our model has better generation 
ability in complex dance synthesis and can achieve fine-grained control through the root trajectory and velocity factor constraints.

{\flushleft\textbf{Experiments on cyclic motion dataset.}}
We select 97,123 frames related to walking and running from the dataset used by Harvey. 
To make a fair comparison, we train our model and Harvey's method~\cite{RobustTweening20} under the same environments using the selected motion samples. 
The generated results are illustrated in Figure~\ref{fig:CompareWalk}, as well as in the supplementary video. 
Since Harvey's method does not control the global position, we modify the root joint's position and rotation
of the synthesized results to avoid the influence of global information on the visual effect. 
From Figure~\ref{fig:CompareWalk} and the supplementary video, we observe that Harvey's method can generate natural 
and smooth motion transitions between keyframes for running motion. 
In contrast, our method has achieved comparable performance to Harvey's method on the cyclic motion dataset from the visual effect. 
However, when it comes to the challenging non-cyclic dance synthesis, our model is far better than Harvey's method 
in the naturalness and variety of dance movements, as shown in Figure~\ref{fig:Compare} and the supplementary video. 
The experimental results validate that our model can not only control complex non-cyclic dance generation, 
but is also suitable for simple cyclic locomotion synthesis. 
Therefore, our approach demonstrates good robustness on a variety of datasets.

\begin{figure}[htbp]
\centering
\includegraphics[width=1.0\linewidth, height=.6\linewidth]{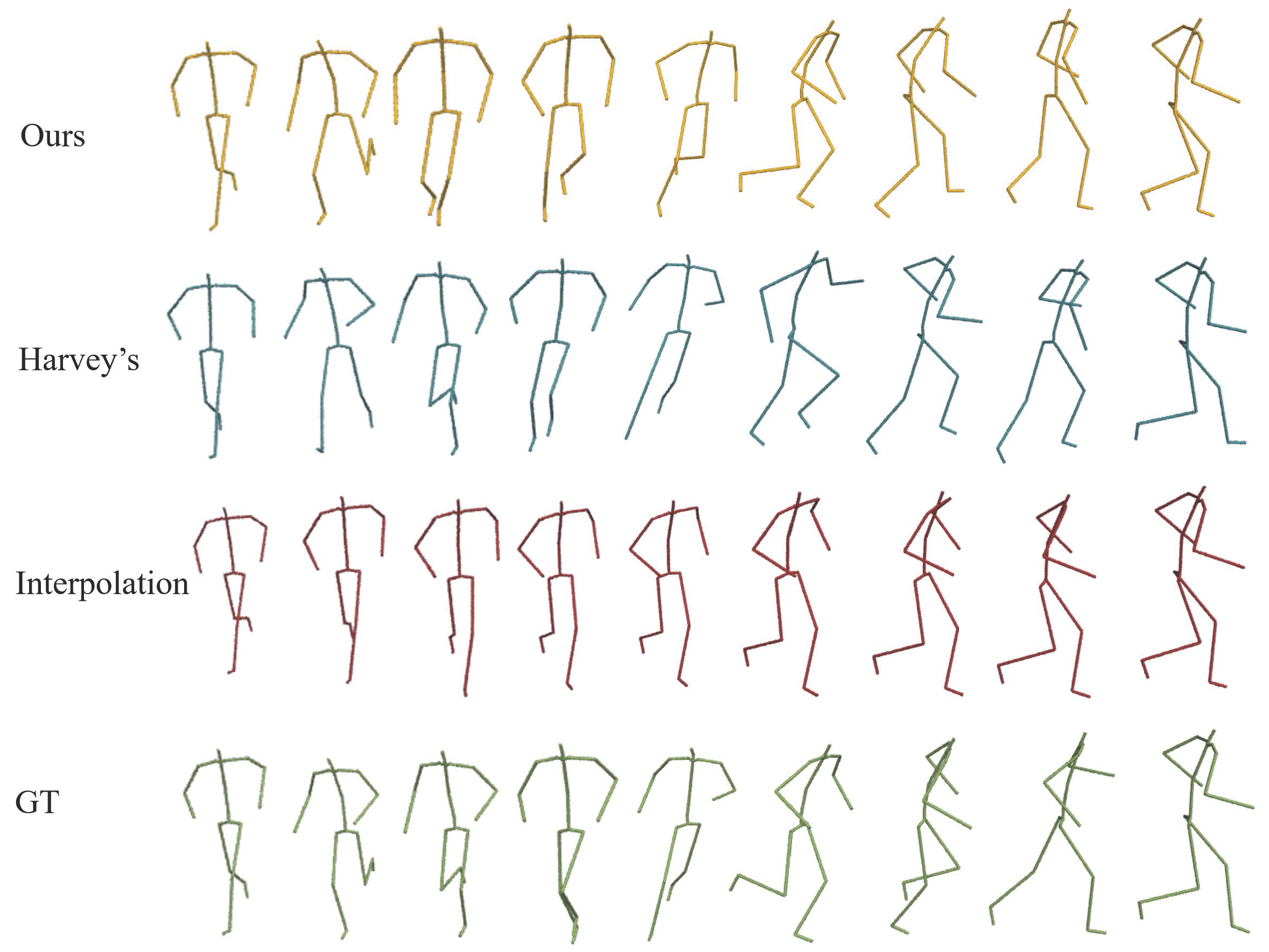}\\
\caption{\label{fig:CompareWalk}%
           Visualization of the generated running sequences by different methods and the ground truth.}
\vspace{-3mm}
\end{figure}

\section{Limitations and discussions}
Our method can achieve fine-grained control of complex movements, but there are still some limitations. 

Firstly, the generated dance motion have the footstep floating problem.
We introduce the foot contact loss in Eq. 7 to constrain footsteps, which is inspired by [WHSZ21]. 
The authors have verified its effectiveness on periodic simple actions, which is consistent with our model on running motion. 
However, dance movements are more complicated compared with cyclic motion, 
Eq. 7 can only alleviate the foot skating to some extent, it hardly solves this problem entirely. 
Besides, since there are many loss terms in our network, and the foot contact labels and 
footstep speeds in Eq. 7 appear in multiple loss terms, the network needs to achieve equilibrium among all the loss terms. 
Therefore, the importance of Eq. 7 may be reduced. Thus, the final generated dance sequences have a footstep floating problem. 
At present, the most effective method to the problem is to use some post-processing methods, 
such as IK~\cite{IK18} or refer to Harvey's method to use GAN~\cite{RobustTweening20}, which will be the focus of our future work.

Secondly, when providing control conditions that differ significantly from the training set, 
our method produces poor results similar to many data-driven tasks. This problem can be avoided by 
expanding the dataset or restricting the control conditions.

Last but not least, since our model has many control constraints, in rare cases, the user may give some 
extremely contradictory control conditions, which will also cause the unreality of the results.
In the future, the relationship between different control conditions can be explored to generate more realistic actions.

\section{Conclusions}
In this paper, we propose a complex motion generation network from keyframes, 
which can realize fine-grained control for complex motion through the trajectory sequence of the root joint 
and the velocity factors of different body parts.
LSTM and Transformer are combined to handle control conditions 
so that our method is able to generate motion sequences in variable length.
We have carried out a large number of comparative experiments to demonstrate the results 
under different constraints. Meanwhile, we have conducted three categories of quantitative evaluation 
and ablation experimental analysis, and have compared our algorithm with the 
state-of-the-art methods, which proves the superiority of our network. 
In the future, we will focus on the footstep floating problem and multiple constraint contradiction issue in complex motion synthesis.
The source code will be released after the paper is accepted. [\url{https://github.com/godzillalla/Dance-Synthesis-Project}].
{\flushleft\textbf{Acknowledgments.} This work was supported in part
by National Key R\&D Program of China (NO.2018YFC0115102), National Natural Science Foundation of China
(No.61872020, U20A20195), Beijing Natural Science Foundation Haidian Primitive Innovation Joint Fund (L182016),
Shenzhen Research Institute of Big Data, Shenzhen, 518000, China Postdoctoral Science Foundation (2020M682827),
Baidu academic collaboration program, and Global Visiting Fellowship of Bournemouth University.}

\bibliographystyle{eg-alpha-doi} 
\bibliography{egbibsample}

\newcommand{\etalchar}[1]{$^{#1}$}
\begin{thebibliography}{\uppercase{HHW{\etalchar{*}}21}}

\bibitem[ALCS18]{IK18}
\textsc{Aristidou A., Lasenby J., Chrysanthou Y., Shamir A.}:
\newblock Inverse kinematics techniques in computer graphics: {A} survey.
\newblock \emph{Computer Graphics Forum 37}, 6 (2018), 35--58.

\bibitem[AZS{\etalchar{*}}17]{dataset2}
\textsc{Aristidou A., Zeng Q., Stavrakis E., Yin K., Cohen{-}Or D., Chrysanthou
  Y., Chen B.}:
\newblock Emotion control of unstructured dance movements.
\newblock In \emph{Proceedings of the {ACM} {SIGGRAPH} / Eurographics Symposium
  on Computer Animation} (Los Angeles, CA, USA, 2017), Eurographics Association
  / {ACM}, pp.~9:1--9:10.

\bibitem[BDK20]{NMT20}
\textsc{Banar N., Daelemans W., Kestemont M.}:
\newblock Character-level transformer-based neural machine translation.
\newblock In \emph{International Conference on Natural Language Processing and
  Information Retrieval} (Seoul, Republic of Korea, 2020), {ACM}, pp.~149--156.

\bibitem[CH07]{MAP07}
\textsc{Chai J., Hodgins J.~K.}:
\newblock Constraint-based motion optimization using a statistical dynamic
  model.
\newblock \emph{ACM Transactions on Graphics 26}, 3 (2007), 8.

\bibitem[CMS{\etalchar{*}}20]{transOD20}
\textsc{Carion N., Massa F., Synnaeve G., Usunier N., Kirillov A., Zagoruyko
  S.}:
\newblock End-to-end object detection with transformers.
\newblock In \emph{European Conference on Computer Vision} (Glasgow, UK, 2020),
  Vedaldi A., Bischof H., Brox T., Frahm J., (Eds.), vol.~12346, Springer,
  pp.~213--229.

\bibitem[dat]{dataset1}
\url{http://dancedb.eu/main/performances}.

\bibitem[ECC{\etalchar{*}}20]{TimeWarping20}
\textsc{Eom H., Choi B., Cho K., Jung S., Hong S., Noh J.}:
\newblock Synthesizing character animation with smoothly decomposed motion
  layers.
\newblock \emph{Computer Graphics Forum 39}, 1 (2020), 595--606.

\bibitem[FCG{\etalchar{*}}21]{GCNdance21}
\textsc{Ferreira J. P.~M., Coutinho T.~M., Gomes T.~L., Neto J.~F., Azevedo R.,
  Martins R., Nascimento E.~R.}:
\newblock Learning to dance: {A} graph convolutional adversarial network to
  generate realistic dance motions from audio.
\newblock \emph{Computers and Graphics 94} (2021), 11--21.

\bibitem[GLSR19]{NPB19}
\textsc{Gaisbauer F., Lehwald J., Sprenger J., Rukzio E.}:
\newblock Natural posture blending using deep neural networks.
\newblock In \emph{Motion, Interaction and Games} (Newcastle upon Tyne, UK,
  2019), Shum H. P.~H., Ho E. S.~L., Cani M., Popa T., Holden D., Wang H.,
  (Eds.), {ACM}, pp.~2:1--2:6.

\bibitem[HHW{\etalchar{*}}21]{huang2021}
\textsc{Huang R., Hu H., Wu W., Sawada K., Zhang M., Jiang D.}:
\newblock Dance revolution: Long-term dance generation with music via
  curriculum learning.
\newblock In \emph{International Conference on Learning Representations}
  (2021).

\bibitem[HP18]{RAT18}
\textsc{Harvey F.~G., Pal C.~J.}:
\newblock Recurrent transition networks for character locomotion.
\newblock In \emph{{SIGGRAPH} Asia 2018 Technical Briefs} (Tokyo, Japan, 2018),
  Zafar N.~B., Zhou K., (Eds.), {ACM}, pp.~4:1--4:4.

\bibitem[HSK16]{HoldenSK16}
\textsc{Holden D., Saito J., Komura T.}:
\newblock A deep learning framework for character motion synthesis and editing.
\newblock \emph{ACM Transactions on Graphics 35}, 4 (2016), 138:1--138:11.

\bibitem[HYNP20]{RobustTweening20}
\textsc{Harvey F.~G., Yurick M., Nowrouzezahrai D., Pal C.~J.}:
\newblock Robust motion in-betweening.
\newblock \emph{ACM Transactions on Graphics 39}, 4 (2020), 60.

\bibitem[HZRS15]{PReLU15}
\textsc{He K., Zhang X., Ren S., Sun J.}:
\newblock Delving deep into rectifiers: Surpassing human-level performance on
  imagenet classification.
\newblock In \emph{{IEEE/CVF} International Conference on Computer Vision}
  (Santiago, Chile, 2015), {IEEE}, pp.~1026--1034.

\bibitem[KB15]{adam}
\textsc{Kingma D.~P., Ba J.}:
\newblock Adam: {a} method for stochastic optimization.
\newblock In \emph{International Conference on Learning Representations (ICLR)}
  (San Diego, CA, USA, 2015).

\bibitem[KGP02]{MotionGraph02}
\textsc{Kovar L., Gleicher M., Pighin F.~H.}:
\newblock Motion graphs.
\newblock \emph{ACM Transactions on Graphics 21}, 3 (2002), 473--482.

\bibitem[KPS03]{motionBeat}
\textsc{Kim T., Park S.~I., Shin S.~Y.}:
\newblock Rhythmic-motion synthesis based on motion-beat analysis.
\newblock \emph{ACM Transactions on Graphics 22}, 3 (2003), 392--401.

\bibitem[LL20]{pose3D2}
\textsc{Lin J., Lee G.~H.}:
\newblock Hdnet: Human depth estimation for multi-person camera-space
  localization.
\newblock In \emph{European Conference on Computer Vision} (Glasgow, UK, 2020),
  Vedaldi A., Bischof H., Brox T., Frahm J., (Eds.), vol.~12363, Springer,
  pp.~633--648.

\bibitem[LLL18]{LeeLL18}
\textsc{Lee K., Lee S., Lee J.}:
\newblock Interactive character animation by learning multi-objective control.
\newblock \emph{ACM Transactions on Graphics 37}, 6 (2018), 180:1--180:10.

\bibitem[LLP13]{MusicSim13}
\textsc{Lee M., Lee K., Park J.}:
\newblock Music similarity-based approach to generating dance motion sequence.
\newblock \emph{Multimedia tools and applications 62}, 3 (2013), 895--912.

\bibitem[LYL{\etalchar{*}}19]{danceToMusic19}
\textsc{Lee H., Yang X., Liu M., Wang T., Lu Y., Yang M., Kautz J.}:
\newblock Dancing to music.
\newblock In \emph{Conference on Neural Information Processing Systems}
  (Vancouver, BC, Canada, 2019), Wallach H.~M., Larochelle H., Beygelzimer A.,
  d'Alch{\'{e}}{-}Buc F., Fox E.~B., Garnett R., (Eds.), pp.~3581--3591.

\bibitem[MBR17]{MartinezB017}
\textsc{Martinez J., Black M.~J., Romero J.}:
\newblock On human motion prediction using recurrent neural networks.
\newblock In \emph{{IEEE} Conference on Computer Vision and Pattern
  Recognition} (Honolulu, HI, USA, 2017), {IEEE} Computer Society,
  pp.~4674--4683.

\bibitem[MCC09]{MAP09}
\textsc{Min J., Chen Y., Chai J.}:
\newblock Interactive generation of human animation with deformable motion
  models.
\newblock \emph{ACM Transactions on Graphics 29}, 1 (2009), 9:1--9:12.

\bibitem[MZZ{\etalchar{*}}19]{NLP19}
\textsc{Ma X., Zhang P., Zhang S., Duan N., Hou Y., Zhou M., Song D.}:
\newblock A tensorized transformer for language modeling.
\newblock In \emph{Conference on Neural Information Processing Systems}
  (Vancouver, BC, Canada, 2019), Wallach H.~M., Larochelle H., Beygelzimer A.,
  d'Alch{\'{e}}{-}Buc F., Fox E.~B., Garnett R., (Eds.), pp.~2229--2239.

\bibitem[RGM19]{STinpaint19}
\textsc{Ruiz A.~H., Gall J., Moreno F.}:
\newblock Human motion prediction via spatio-temporal inpainting.
\newblock In \emph{{IEEE/CVF} International Conference on Computer Vision}
  (Seoul, Korea (South), 2019), {IEEE}, pp.~7133--7142.

\bibitem[SNI06]{DanceAnimation06}
\textsc{Shiratori T., Nakazawa A., Ikeuchi K.}:
\newblock Dancing-to-music character animation.
\newblock \emph{Computer Graphics Forum 25}, 3 (2006), 449--458.

\bibitem[TJM18]{lstmMelody18}
\textsc{Tang T., Jia J., Mao H.}:
\newblock Dance with melody: An lstm-autoencoder approach to music-oriented
  dance synthesis.
\newblock In \emph{{ACM} Multimedia Conference on Multimedia Conference}
  (Seoul, Republic of Korea, 2018), Boll S., Lee K.~M., Luo J., Zhu W., Byun
  H., Chen C.~W., Lienhart R., Mei T., (Eds.), {ACM}, pp.~1598--1606.

\bibitem[VSP{\etalchar{*}}17]{Transformer17}
\textsc{Vaswani A., Shazeer N., Parmar N., Uszkoreit J., Jones L., Gomez A.~N.,
  Kaiser L., Polosukhin I.}:
\newblock Attention is all you need.
\newblock In \emph{Conference on Neural Information Processing Systems} (Long
  Beach, CA, {USA}, 2017), Guyon I., von Luxburg U., Bengio S., Wallach H.~M.,
  Fergus R., Vishwanathan S. V.~N., Garnett R., (Eds.), pp.~5998--6008.

\bibitem[WCX21]{RNN-AT21}
\textsc{Wang Z., Chai J., Xia S.}:
\newblock Combining recurrent neural networks and adversarial training for
  human motion synthesis and control.
\newblock \emph{IEEE Transactions on Visualization and Computer Graphics 27}, 1
  (2021), 14--28.

\bibitem[WFH08]{GPDM08}
\textsc{Wang J.~M., Fleet D.~J., Hertzmann A.}:
\newblock Gaussian process dynamical models for human motion.
\newblock \emph{{IEEE} Transactions on Pattern Analysis and Machine
  Intelligence 30}, 2 (2008), 283--298.

\bibitem[WHSZ21]{STML21}
\textsc{Wang H., Ho E. S.~L., Shum H. P.~H., Zhu Z.}:
\newblock Spatio-temporal manifold learning for human motions via long-horizon
  modeling.
\newblock \emph{IEEE Transactions on Visualization and Computer Graphics 27}, 1
  (2021), 216--227.

\bibitem[YWJ{\etalchar{*}}20]{ChoreoNet20}
\textsc{Ye Z., Wu H., Jia J., Bu Y., Chen W., Meng F., Wang Y.}:
\newblock Choreonet: Towards music to dance synthesis with choreographic action
  unit.
\newblock In \emph{{ACM} Multimedia Conference on Multimedia Conference}
  (Seattle, WA, USA, 2020), Chen C.~W., Cucchiara R., Hua X., Qi G., Ricci E.,
  Zhang Z., Zimmermann R., (Eds.), {ACM}, pp.~744--752.

\bibitem[ZFS{\etalchar{*}}20]{pose3D1}
\textsc{Zhen J., Fang Q., Sun J., Liu W., Jiang W., Bao H., Zhou X.}:
\newblock {SMAP:} single-shot multi-person absolute 3d pose estimation.
\newblock In \emph{European Conference on Computer Vision} (Glasgow, UK, 2020),
  Vedaldi A., Bischof H., Brox T., Frahm J., (Eds.), vol.~12360, Springer,
  pp.~550--566.

\bibitem[ZJJ{\etalchar{*}}20]{PointTrans}
\textsc{Zhao H., Jiang L., Jia J., Torr P. H.~S., Koltun V.}:
\newblock Point transformer.
\newblock \emph{arXiv preprint arXiv:2012.09164} (2020).

\bibitem[ZS09]{MotionGraph09}
\textsc{Zhao L., Safonova A.}:
\newblock Achieving good connectivity in motion graphs.
\newblock \emph{Graphical Models 71}, 4 (2009), 139--152.

\bibitem[ZvdP18]{autocompletion18}
\textsc{Zhang X., van~de Panne M.}:
\newblock Data-driven autocompletion for keyframe animation.
\newblock In \emph{In Proceedings of International Conference on Motion,
  Interaction, and Games} (Limassol, Cyprus, 2018), {ACM}, pp.~10:1--10:11.

\bibitem[ZZT{\etalchar{*}}20]{FPT20}
\textsc{Zhang D., Zhang H., Tang J., Wang M., Hua X., Sun Q.}:
\newblock Feature pyramid transformer.
\newblock In \emph{European Conference on Computer Vision} (Glasgow, UK, 2020),
  Vedaldi A., Bischof H., Brox T., Frahm J., (Eds.), vol.~12373, Springer,
  pp.~323--339.

\end{thebibliography}


\end{document}